\documentclass[12pt]{article} 

\usepackage{amssymb}
\usepackage{amsmath}
\usepackage{graphicx}
\usepackage{subfigure}
\usepackage{makeidx}
\usepackage{multicol}
\usepackage[export]{adjustbox}
\usepackage[justification=centering]{caption}
\usepackage{chngcntr}
\counterwithout{figure}{section}
\usepackage{float}

\usepackage[table,xcdraw]{xcolor}

\usepackage{multirow}
\def\B#1{#1}
\def\G#1{#1}

\usepackage{fancyhdr}
\let\origtitle\title 
\renewcommand{\title}[1]{\lfoot{\textit{#1}}\origtitle{\textbf{#1}}}
\renewcommand{\sectionmark}[1]{\markboth {}{}}

\date{}

\title{Detecting Informative Channels: 
ActionFormer}

\begin{document}
\maketitle
\thispagestyle{fancy}
\centering

\author{Kunpeng Zhao \footnote{zhao.kunpeng214@gmail.com},
Asahi Miyazaki \footnote{miyazaki.asahi676@mail.kyutech.jp},
Tsuyoshi Okita* \footnote{tsuyoshi@ai.kyutech.ac.jp}
}\\
\thanks{$^1$$^2$$^3$Kyushu Institute of Technology}

% \author{Hoang Anh Vy Ngo* \footnote{anhvy3008@gmail.com}, Vu Nguyen Phuong Quynh \footnote{quynhvnp26@gmail.com}, Noriyo Colley \footnote{noriyo@med.hokudai.ac.jp}, Shinji Ninomiya \footnote{s-nino@hs.hirokoku-u.ac.jp}, Satoshi Kanai \footnote{kanai@ssi.ist.hokudai.ac.jp}, Shunsuke Komizunai\footnote{komizunai.shunsuke@kagawa-u.ac.jp}, Atsushi Konno \footnote{konno@ssi.ist.hokudai.ac.jp}, Misuzu Nakamura \footnote{misuzun@jikei.ac.jp}, Sozo Inoue \footnote{sozo@brain.kyutech.ac.jp}}\\
% \thanks{$^1$$^2$$^9$Kyushu Institute of Technology, $^3$$^5$$^7$Hokkaido University, $^4$Hiroshima International University, $^6$Kagawa University, $^8$JIKEI University}

%%%%%%%% abstract from here %%%%%%%%%%
\abstract{
%In this paper, we propose using pose estimation extracted from videos to recognize the nurses’ activity when doing endotracheal suctioning. Endotracheal suctioning is a sophisticated method that is very invasive and may accompany risks for patients. As home healthcare becomes more prevalent, there is an urgent need for more certified individuals who can perform endotracheal suctioning. However, quantitative research on nurse care activity recognition in endotracheal suctioning is still limited. To address this issue, our study aims to recognize 9 suctioning activities from video recordings by extracting their pose estimation to classify activities. The videos recorded 10 nurses and 12 nursing students performing tracheal suctioning on the simulation system. Because the videos were taken in a real-world environment, there are some obstacles to overcome such as people in the background, and nurses standing out of the frame. Therefore, we proposed an algorithm to define and track the main subject. Also, the missing keypoints problems due to the performance of the pose estimation algorithm are improved by smoothing keypoints. After that, these key points are input for Random Forest model to classify activities. By comparing to the result of baseline approach without post-processing steps, we found that applying post-processing steps after pose estimation can improve the performance of our learning model. Our proposed method achieved an accuracy of 54\% and F1-score of 46\%, while the baseline approach's accuracy is 51\% and F1 score is 42\%.
Human Activity Recognition (HAR) has recently witnessed advancements with Transformer-based models. Especially, ActionFormer shows us a new perspectives for HAR in the sense that this approach gives us additional outputs which detect the border of the activities as well as the activity labels. ActionFormer was originally proposed with its input as image/video. However, this was converted to with its input as sensor signals as well. 
We analyze this extensively in terms of deep learning architectures. 
Based on the report of high temporal dynamics which limits the model's ability to capture subtle changes effectively and of the interdependencies between the spatial and temporal features. We propose the modified ActionFormer which will decrease these defects for sensor signals. The key to our approach lies in accordance with the Sequence-and-Excitation strategy to minimize the increase in additional parameters and opt for the swish activation function to retain the information about direction in the negative range. Experiments on the WEAR dataset show that our method achieves substantial improvement of a 16.01\% in terms of average mAP for inertial data.
%In this paper, we present a novel channel-wise enhancement framework for activity recognition models. While existing works often focus on extracting higher-level features using CNN or RNN architectures for human activity recognition (HAR) tasks, our approach leverages the Transformer framework by introducing a Channel-wise Enhancement Module. This module is specifically designed to direct the model's focus toward informative channels, thereby facilitating more effective extraction of spatial and temporal features. Extensive experiments on the WEAR dataset demonstrate that our method achieves substantial improvement over the ActionFormer baseline. Co-\newline mpared to existing models on the WEAR dataset, \B{we showed that
%our method outperforms the previous ActionFormer equivalent.}
%also achieves the 
%best 
%considerable performance. 
}

\section{Introduction}

 Human activity recognition (HAR) using sensor signals \cite{Cleland2024,Zolfaghari2024,Phan2024,Bjorn2022}
 has garnered significant attention due to its wide-ranging applications in healthcare, sports monitoring, and wearable technologies. 
 Recent advances in deep learning have revolutionized the field, enabling effective processing of temporal dynamic sequences through automated feature extraction. 
 In the last decade, Transformer models have demonstrated remarkable capabilities across various tasks. This success can be attributed to their powerful sequence modeling capabilities by the internal attention mechanism \cite{attention}. A representative architecture among them is ActionFormer \cite{zhang2022actionformer} in temporal action localization (TAL) tasks, which is a Transformer architecture model to identify actions and locate their start and end times from video data. Bock et al. \cite{bock2023wear} extended its application to activity localization HAR tasks, utilizing sensor signals as input to predict activity labels and locate activity boundries. 

% However, we observe that sensor signals exhibit high temporal dynamics, which limits the model's ability to capture subtle changes effectively. 

However, despite the success of existing approaches, several critical gaps remain unaddressed. First, sensor signals exhibit high temporal dynamics, which limits the model's ability to capture subtle changes effectively. Second, existing methods typically rely on sptial and temporal feature extractions. For instance, \cite{szegedy2015going} strengthens the representations generated by CNNs by incorporating the 'Inception module', while \cite{bn} introduces batch normalization to adapt the new distribution between layers. Further, studies such as \cite{bell2015ion} and \cite{newell2016stackedhourglassnetworkshuman} emphasize multi-scale representations to enable models to capture diverse spatial relationships, while \cite{jaderberg2016spatialtransformernetworks} integrates spatial attention into the model's architecture. These approaches have demonstrated their effectiveness in capturing spatial correlations among features.
Interestingly, \cite{hu2018squeeze} approaches this problem from the perspective of channel relationships, enhancing the representational power of CNNs. It enables models to recalibrate features by emphasizing critical ones. In HAR tasks, similar efforts have been made to focus on channel relationships to enhance the task performance.
% To enhance HAR tasks, 
\cite{CrossHAR} and \cite{hopp2024ta} normalize each of the inertial axes separately, which is a channel-wise normalization method. The authors \cite{CrossHAR} argue that avoiding the destruction of channels-specific data is beneficial for the HAR tasks.
\B{These methods may inadvertently suppress critical channel-specific information, leading to suboptimal performance in complex HAR scenarios.}

\G{At the beginning, we assumed in this paper that we look at each dimension of the representation of the sensor signals (after feature extraction or encoding from the input signals) as a channel. Features at each dimension would have importance, in terms of temporal dynamics and temporal-spatial relationships, with regard to the classification of HAR task. This is our starting point in this paper.}

\G{Thus, the research questions in this paper are the following:}

\begin{itemize}
\item  \G{In order to capture the temporal dynamics of sensor data, this may be appeared in the local channels rather than the global channels. Then, these can be handled by the importance-based feature extactor.}
\item  \G{In order to assess the relationships between the temporal feature and the spatial feature, these correlations may be appeared in the time-series analysis of importance in each channel.}
\end{itemize}

\G{By these questions, we start from the squeeze-and-excitation module \cite{hu2018squeeze} which considers the similar situations by weighting the importance of channels after convolutional networks, together with swish activation function \cite{Swish}. Then, we modify this to form the channel-wise enhancement module.}

%\B{To address these gaps, we propose a novel channel-wise enhancement framework for HAR tasks. Our work is driven by the following research questions:}
%\begin{itemize}
%    \item \B{How can we effectively leverage channel-wise relationships in sensor data to improve the extraction of spatial and temporal features?}
%    \item \B{Can a lightweight channel-wise enhancement module be integrated into Transformer-based models to enhance their performance without significant computational overhead?}
%    \item \B{How does the channel-wise framework perform compared to state-of-the-art methods in complex HAR scenarios?}
%\end{itemize}

\G{The contribution of this work consists of the following:}
\begin{itemize}
\item \G{We propose the method within the signal version of ActionFormer to weigh the importance of each channel in the representation of signals, which we name the adaptive channel-wise enhancement module. }
\item \G{We showed that this module made improvements on WEAR dataset over several baseline methods.}
\end{itemize}

\begin{figure}
    %\centering
    \includegraphics[width=\linewidth]{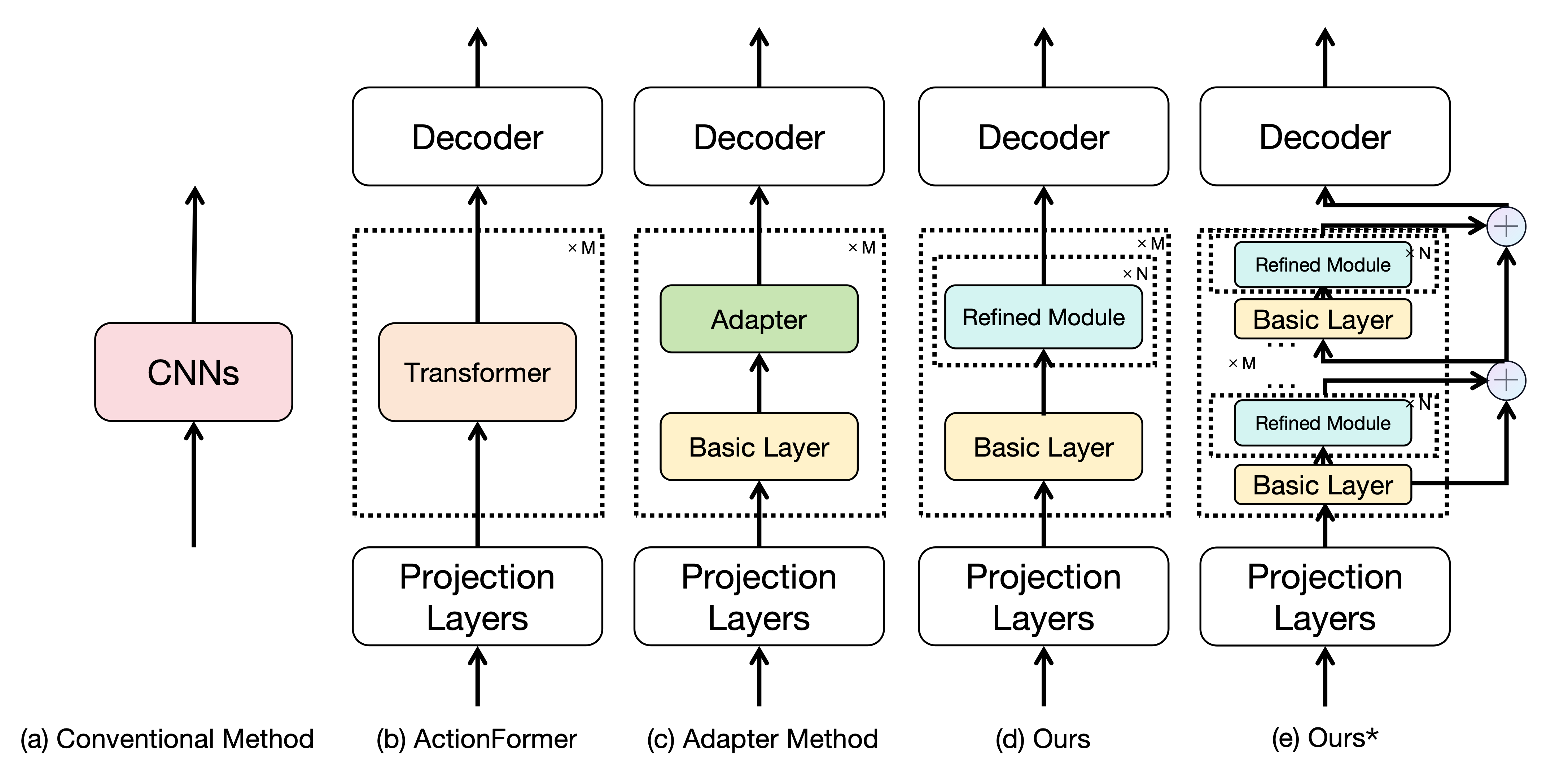}
    \caption{
    % Architectural comparison of conventional HAR frameworks and our proposed approaches. (a) Conventional HAR architecture: The raw inertial data is initially processed through CNNs or RNNs in the projection layer for feature extraction, followed by a backbone network composed of basic layers for feature processing. (b) ActionFormer architecture: An enhanced framework that employs transformer-based layers as the primary backbone components. (c) Our proposed framework: Integrates dedicated channel-wise enhancement modules following each basic layer to strengthen channel-specific feature representations. (d) Alternative module placement strategy: An optimized configuration designed to minimize information loss while reducing computational complexity and memory requirements.
    \B{Architectural comparison of conventional HAR frameworks and our proposed approaches. (a) Conventional method: CNNs. (b) ActionFormer architecture: An enhanced framework that employs Transformer-based layers as the primary backbone components. (c) Adapter method (Standard Adapter and TIA methods): Use adapter after basic layer to strengthen the framework. (d) Our proposed framework: Integrates dedicated 
    % channel-wise enhancement
    modules following each basic layer to strengthen 
    % channel-specific 
    feature representations. (e) Alternative module placement strategy: An optimized configuration designed to minimize information loss while reducing computational complexity and memory requirements.}
    }
    \label{fig:methods}
\end{figure}

\section{Related Work}

\B{\subsection{Time Series Algorithms}}
Deep learning methods have demonstrated significant success in processing temporal dynamic sequences through their advanced pattern recognition capabilities and automated feature extraction using neural network architectures. The development of sequential data processing models can be broadly categorized into three main approaches:
The first category comprises CNN-based models \cite{van2016wavenet, bai2018empirical, chen2020probabilistic}. These models excel at capturing short-term dependencies by leveraging convolution and pooling operations to automatically learn complex patterns and relationships within temporal sequences.
The second category includes RNN-based models \cite{RNN, LSTM, GRU}, which are specifically designed to handle variable-length sequential data. These models incorporate specialized memory units and gating mechanisms, such as Long Short Term Memory (LSTM) \cite{LSTM} and Gated Recurrent Units (GRU) \cite{GRU}, enabling effective capture of temporal dependencies.
The third category consists of Transformer-based models \cite{attention, lim2021temporal, li2019enhancing, kitaev2020reformer, zhou2021informer}. These models employ attention mechanisms to identify and focus on the most relevant components of sequential data, thereby enhancing the model's ability to extract valuable information.

% \textbf{ActionFormer \& WEAR.}
% In the context of HAR, Transformers have shown particularly promising results. A notable example is ActionFormer \cite{zhang2022actionformer}, one of the pioneering transformer-based models for TAL tasks. It combines multiscale feature representation with local self-attention and employs a lightweight decoder for temporal moment classification and action boundary estimation. Building upon this work, Bock et al. \cite{bock2023wear} introduced the WEAR dataset, an outdoor dataset comprising untrimmed inertial (acceleration) and egocentric video data from 18 participants performing 18 distinct workout activities across 10 different outdoor locations. They integrated both ActionFormer \cite{zhang2022actionformer} and TriDet \cite{Shi_2023_CVPR} into the WEAR library for evaluation purposes. More recently, Hartleb et al. \cite{hopp2024ta} enhanced performance on this dataset by incorporating adapter modules into the TriDet framework and implementing data augmentation techniques.

\B{\subsection{Spatial and Temporal Dependencies}}
Beyond architectural innovations, significant research effort has focused on improving spatial and temporal feature extraction capabilities. Several key developments have emerged in this area: the Inception module \cite{szegedy2015going} and batch normalization \cite{bn} have enhanced feature representations; multi-scale representations \cite{bell2015ion, newell2016stackedhourglassnetworkshuman} and spatial attention mechanisms \cite{jaderberg2016spatialtransformernetworks} have improved the capture of diverse spatial relationships; and channel relationship methods, exemplified by SE networks \cite{hu2018squeeze}, have enabled feature recalibration by emphasizing critical features. Specifically in HAR tasks, channel-wise normalization techniques \cite{CrossHAR} \cite{hopp2024ta} have further improved performance by preserving channel-specific characteristics.

\B{\subsection{Squeeze-and-Excitation Strategy}} 
%The Squeeze-and-Excitation(SE) 
The SE network \cite{hu2018squeeze} has been shown in previous research\cite{zhang2021spatiotemporal, an2018squeeze, chen2020fused, chen2020esenet, wang2021action, zhang2023temporal, hua2022drn, kapoor2023enhancing, cao2019gcnet, tan2020efficientnetrethinkingmodelscaling} to effectively enhance the performance of models in action recognition tasks. An et al. \cite{an2018squeeze} utilized the SE \B{\cite{hu2018squeeze}} module in combination with both ResNet \cite{he2016deep} and LSTM \cite{hochreiter1997long} networks to achieve the recalibration of spatial and temporal features, respectively. Their proposed SE-LRCN model \B{\cite{an2018squeeze}} considers the significance of both spatial and temporal features, 
% effectively uncovering the intrinsic dependencies at pixel and frame granularity in video actions. 
achieved competitive results in the HMDB51 \cite{kuehne2011hmdb} and UCF101 \cite{soomro2012ucf101} action recognition benchmarks. 
\cite{chen2020fused, wang2021action} have introduced the SE or SE-based modules into 2DCNN networks to model the relationship between spatial feature channels.
Chen et al. \cite{chen2020esenet} further investigated by simply integrating 3DSE (a 3d version of the SE module) with the 3DCNN network, and the proposed ESENet \B{\cite{chen2020esenet}} network outperformed the baseline on the video action recognition task. There has also been some work focused on designing improved SE module for specific activity recognition tasks, such as \cite{zhang2023temporal, shu2022expansion}. Overall, SE \B{\cite{hu2018squeeze}} has been proven by extensive research to work well in helping with activity recognition tasks.

\section{Methodology}
\label{section:Method}
This section introduces our channel-wise enhancement method for ActionFormer \cite{zhang2022actionformer}. 
We assume that ActionFormer is 
given the inertial features extracted from the original signal data using DeepConvLSTM \cite{deepconvlstm} or Attend-and-Discriminate \cite{abedin2021attend}. These features form the {\sl channels} from now on.
First, we define the notations used in the activity localization HAR tasks. Next, we elaborate on our method, which is specifically tailored for inertial data. Subsequently, we introduce the enhanced ActionFormer model, improved using our proposed method. %In addition, we propose an optional approach specifically designed for visual data. 
Finally, we discuss alternative connection methods for the proposed modules.

\subsection{Notations}
% HAR typically involves one of the following two main task settings. The first is a classification task, where the goal is to predict the activity label for a given segment of sensor data. The second is a detection task, where both the activity category and the start and end times of the activity need to be predicted. Our solution adopts the second task design and uses the extracted features as input, the whole task can be formulated as follows:

% Given input feature data $X$: $X = [\mathbf{x}_1, \mathbf{x}_2, \dots, \mathbf{x}_T]$, where $T$ is the length of the time series and each input sample $\mathbf{x}_i$ may be features extracted from the sensor data. Output $\hat{A} = \{(s_i, e_i, c_i)\}_{i=1}^N$, where $N$ is the number of detected activities, $(s_i, e_i)$ are the start and end times of the i-th activity, and $c_i \in \{1, 2, \dots, C\}$ is the activity categories, where $C$ is the total number of ground truth activites. 
The activity localization HAR task is defined as predicting the activity category along with its start and end times from temporal sequential data. The task can be formulated as follows:

Given input feature data $X$: $X = [\mathbf{x}_1, \mathbf{x}_2, \dots, \mathbf{x}_T]$, where $T$ is the length of the time series and each input sample $\mathbf{x}_i$ represents the extracted features from raw sequences. Output is defined as $\hat{A} = \{(s_i, e_i, c_i)\}_{i=1}^N$, where $N$ is the number of detected activities, $(s_i, e_i)$ are the start and end times of the i-th activity, and $c_i \in \{1, 2, \dots, C\}$ denotes the activity categories, where $C$ is the total number of ground truth activites. 

\begin{figure}
    \centering
    \includegraphics[width=0.5\linewidth]{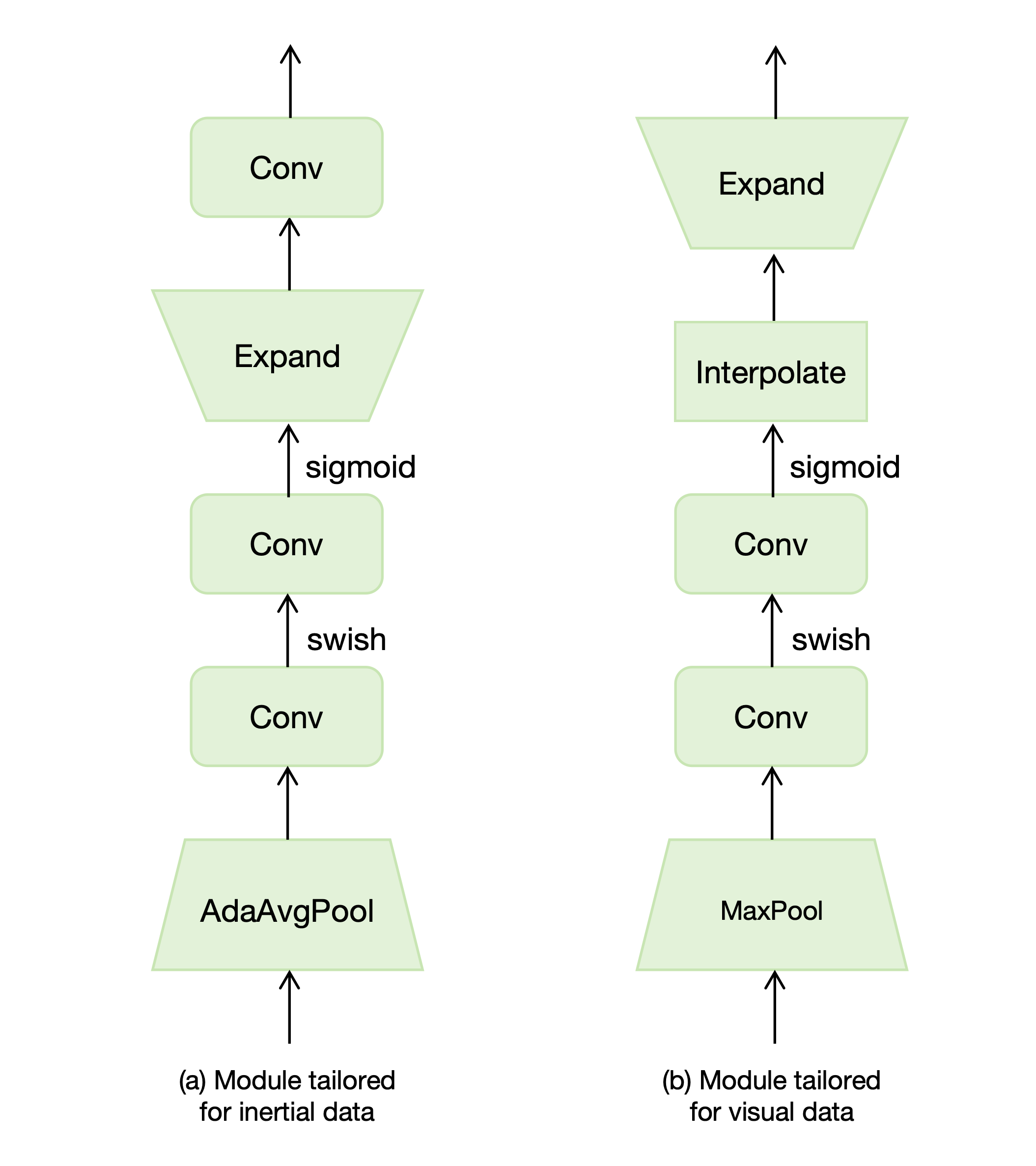}
    \caption{\B{(a)}Adaptive Channel-wise Enhancement Module tailored for inertial data versus; \B{(b)} MaxPool Chanel-wise Enhancement Module tailored for visual data}
    \label{fig:Two}
\end{figure}

\subsection{Adaptive Channel-wise Enhancement}
Sensor signals are typically characterized as low-dimensional time series, with critical features predominantly distributed across specific channels. Therefore, we propose compressing the input inertial sequences into channel-wise representations. To this end, we consider utilizing PyTorch's $AdaptiveAvgPool1d(1)$ function to perform one-dimensional adaptive average pooling on the input signal. This approach ensures that the number of feature channels remains unchanged while specifying the output size as 1.
\B{To be precise, the original data shape is [batch size $\times$ feature channels $\times$ sequence length], and after $AdaptiveAvgPool1d(1)$, the shape becomes [batch size $\times$ feature channels $\times$ 1], thus achieving the purpose of retaining only channel information.}
By computing the importance of each channel and applying scaling factors accordingly, the method amplifies salient channel information, thereby enhancing the overall feature representations. The module structure is illustrated in Figure \ref{fig:Two} (a), and the mathematical formulation is described as follows:
Given the inertial feature input $\mathbf{X} \in \mathbf{R}^{B \times C \times T}
$, where $B$ denotes batch size, $C$ represents feature channels and $T$ corresponds to the sequence length. The core formula for channel weighting is defined as follows:
\begin{equation}
\label{e1}
	\mathbf{W}_{\text{C}} = \text{sigmoid}\left(\text{conv}_{\text{2}}\left(\text{swish}\left(\text{conv}_{\text{1}}\left(\text{avgPool}(\mathbf{X})\right)\right)\right)\right)
\end{equation}
The input features will be weighted using this weight and the output will be processed through an  $1 \times 1$ convolutional layer:
\begin{equation}
	\mathbf{Y} = \text{conv1d}_{\text{$1 \times 1$}}\left(\mathbf{X} \odot \mathbf{W}_{\text{C}}\right)
\end{equation}

In the design of our method, we employ the Squeeze-and-Excitation strategy \cite{hu2018squeeze} to minimize computational overhead while avoiding excessive redundant information. In our experiments, we set the reduction factor to 16, though exploring optimal reduction factors for different inputs remains an essential step. Regarding the choice of activation function, we opt for swish \cite{Swish} over ReLU \cite{relu} due to its ability to introduce nonlinearity while maintaining gradients in negative regions, thereby mitigating the vanishing gradient problem. 
\B{Specifically, in sensor signals, many features may still have physical meanings in the negative range, such as the directional information of acceleration or angular velocity. Therefore, we aim to select an activation function that can retain the useful information in the negative range. The characteristic of the swish \cite{Swish} activation function is similar to ReLU \cite{relu}, but does not entirely zero out negative values. Instead, it gradually decreases in the negative region, aligning well with our expectations, which meets our expectations.}
The swish activation function is defined as:
\begin{equation}
	\text{swish}(x) = x \cdot \sigma(\beta x)
\end{equation}
where $\sigma(x) = \frac{1}{1 + e^{-x}}$ represents the sigmoid function. We evaluate the performance of the swish function with various $\beta$ values $\left[ 0.1, 0.5, 1.0, 5, 10 \right]$
 and set $\beta$ as a learnable parameter, finalize to use the swish function with a $\beta$ value of 1.0 based on the test results, namely the SiLU.

%\subsection{Baseline Election and The Improved Model}
\subsection{\B{Proposed Model}}

\begin{figure*}[!ht]
    \centering
    \includegraphics[width=\linewidth]{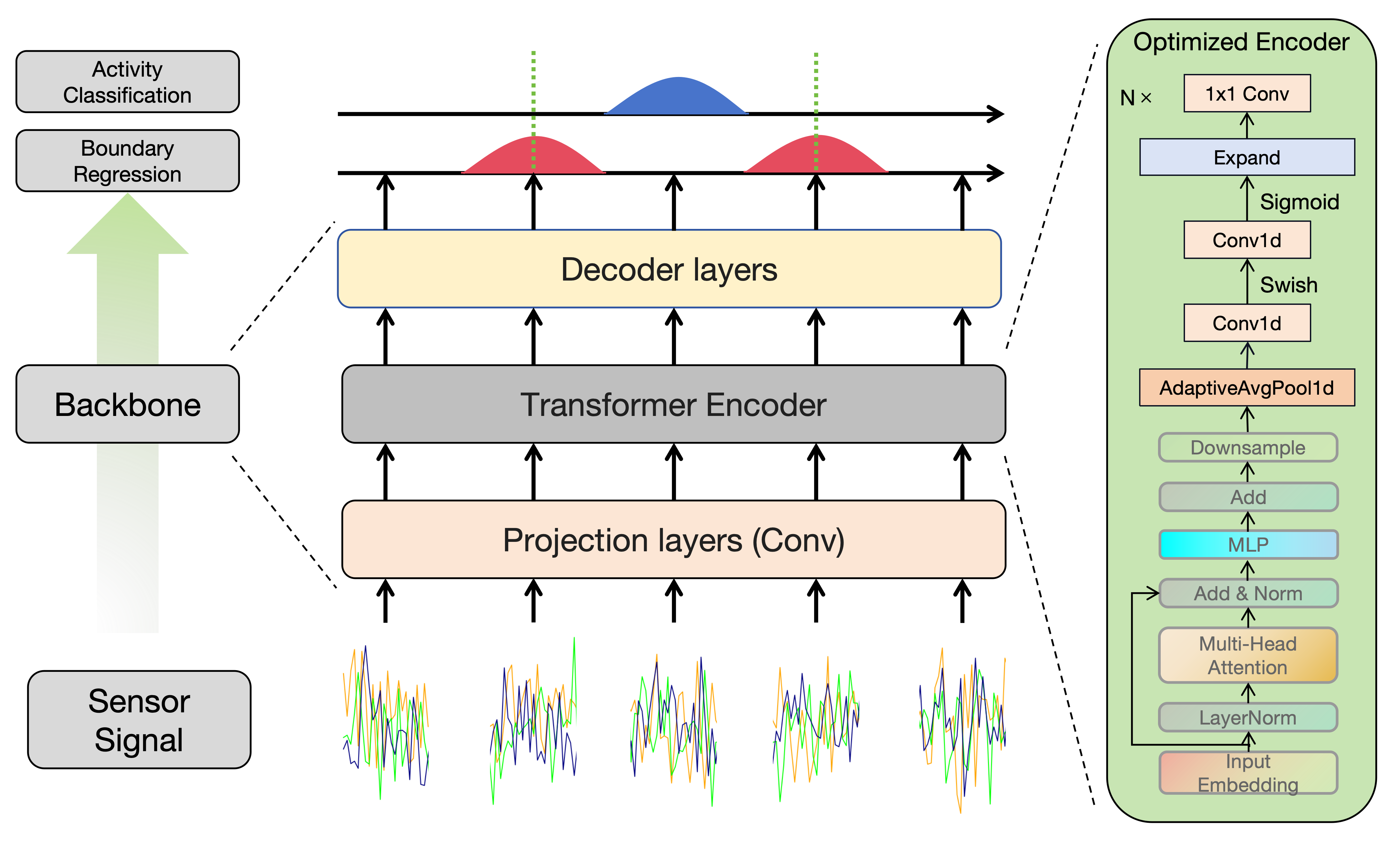}
    \caption{Overview of our backbone. We build our refined encoder (illustrated within the green box on the right side of the figure), which can direct the model's attention toward important channels, thereby improving the ability to extract spatial and temporal features.}
    \label{fig:backbone}
\end{figure*}

Figure \ref{fig:backbone} illustrates the overall architecture of our improved ActionFormer model, which has been enhanced for channel-wise information extraction. 
Specifically, we enhance its capability to model inertial sequences using the Adaptive Channel-wise Enhancement module. Our modifications are implemented based on the codebase provided by bock et al \cite{bock2023wear}.

Our proposed model closely follows the ActionFormer \cite{zhang2022actionformer}, with channel-wise enhancements. The initial stage comprises two one-dimensional convolutional layers that function as projection layers. These layers map the input signal data $X$ into a D-dimensional vector space, with ReLU serving as the activation function. The inclusion of projection layers preceding the transformer network facilitates the processing of time series data \cite{li2019enhancing}.
Subsequently, the embedded vector is fed into the transformer encoder. 
% The encoder, as provided by the Bock et al. \cite{bock2023wear}, comprises 7 transformer blocks. 

Each transformer block includes layer normalization, a multi-head self-attention mechanism, a multilayer perceptron (MLP), and a downsampling technique. In the encoder component, the original model employs a stack of seven transformer blocks to generate a feature pyramid.

Our contribution lies in optimizing the encoder of the transformer. Specifically, we enhance its capability to model inertial sequences using the adaptive channel-wise enhancement module. 
% The figure only illustrates the first strategy, which uses inertial data as input.

The final component is a lightweight decoder that processes the generated feature pyramid and outputs the label for the input signal data. This decoder is implemented as a convolutional network with dual heads: a classification head and a regression head. The classification head is responsible for predicting the labels of the signal data, while the regression head estimates the distances to the onset and offset of the activities.

\subsection{MaxPool Chanel-wise Enhancement}
In this work, we primarily focus on enhancing ActionFormer's ability to extract spatiotemporal features from sensor inputs, achieving significant improvements. However, when testing the model with video inputs using this approach, its performance showed a decline. To address this, we also propose a method tailored to improve the model's performance specifically for video inputs, as illustrated in Figure \ref{fig:Two} (b).  We assume that only certain information between adjacent video frames is relevant to the activity context, while other aspects, such as background information, remain highly similar. To filter out this redundant information, we replace the adaptive average pooling layer with a max pooling layer. Moreover, we employ interpolation to restore the original sequence length before feature scaling. This module can be described in the following formulation.
% \[
% \mathbf{X}_{\text{out}} = \mathbf{X} \odot \text{Interpolate} \left( 
% \sigma \left( \text{Swish} \left( \mathbf{X}_{\text{max}} \cdot \mathbf{W}_1 + \mathbf{b}_1 \right) \cdot \mathbf{W}_2 + \mathbf{b}_2 \right), \, \text{size}=T \right),
% \]

Assume the input video feature $\mathbf{X} \in \mathbf{R}^{B \times C \times T}$, where $B$ denotes the batch size, $C$ represents feature channels, and $T$ corresponds the sequence length. After applying local max pooling, we obtain $\mathbf{X}_{\text{max}}$:

\begin{equation}
\mathbf{X}_{\text{max}} = \text{MaxPool1D}(\mathbf{X}, k, s)
\end{equation}

We replace $\text{AvgPool}(\mathbf{X})$ in Equation \ref{e1} with $\mathbf{X}_{\text{max}}$ and perform the same operation.
The $\mathbf{W}_{\text{C}}$ are then interpolated back to the original sequence length $T$:

\begin{equation}
\mathbf{Y}_{\text{interp}} = \text{Interpolate}(\mathbf{W}_{\text{C}}, T)
\end{equation}

Finally, the output feature is computed as:

\begin{equation}
\mathbf{X}_{\text{out}} = \mathbf{X} \odot \mathbf{Y}_{\text{interp}}
\end{equation}

Note that the kernel size of max pooling layer is set to 3 and stride is set to 2, the most optimal settings still require further exploration based on the specific datasets.

\subsection{Alternative Connections}

In our model, the basic layer consists of seven layers, categorizing it as a shallow network structure. In contrast, many works have opted for deeper network structures to extract higher-level features and improve model performance. However, deeper networks often suffer from issues such as gradient vanishing. A classic solution to this problem is the ResNet architecture proposed by He et al. \cite{he2016deep}, which directly fuses the outputs from shallow and deep layers.
To enable our module to adapt to deeper network structures, we have designed a cross-layer connection mechanism, as illustrated in Figure \ref{fig:methods} (d). By directly passing information between layers, this mechanism facilitates better fusion of low-level and high-level features, enhancing the model's capacity to learn from complex data representations. 
% To distinguish the different variants, we name the standard adaption design as CE-HAR, and the alternative placement as CE-HAR*.

\section{Experiments}
% \textbf{Dataset}.
% We conduct the experiments on the WEAR dataset  \cite{bock2023wear}, an outdoor sports dataset for both vision- and inertial-based human activity recognition.
% We use the inertial data provided by WEAR for the main experiments.

% Compared to 
% %activity recognition 
% action recognition datasets such as THUMOS14 \cite{idrees2017thumos}, ActivityNet-1.3 \cite{caba2015activitynet} and EPIC-Kitchens 100 \cite{dima2020rescaling}, in addition to video data, the WEAR dataset \cite{bock2023wear} also provides separate inertial data collected from sensors. 
\subsection{\B{WEAR Dataset Description}}
\B{
The WEAR (An Outdoor Sports Dataset for Wearable and Egocentric Activity Recognition) dataset is a benchmark dataset designed for HAR, focusing on outdoor sports activities using wearable inertial sensors and egocentric vision. The dataset aims to provide a realistic and challenging HAR scenario by incorporating continuous untrimmed activity sequences, multiple recording locations, and diverse movement patterns.
}

\subsubsection{\B{Data Collection Process}}
\B{
\begin{itemize}
\item \textbf{Participants:} 18 individuals (10 male, 8 female), aged 28 years on average ($\pm$5), participated in the study.
\item \textbf{Recording Locations:} Data was collected at 10 different outdoor locations under varying weather conditions to ensure dataset diversity.
\item \textbf{Sensor Setup:}
\begin{itemize}
\item \textbf{IMU Sensors:} Four Bangle.js smartwatches were placed on both wrists and ankles to capture 3D acceleration data at 50Hz with a sensitivity of $\pm$8g.
\item \textbf{Egocentric Video:} A GoPro Hero 8 camera, mounted on the participant’s head, recorded 1080p video at 60 FPS with a 45-degree downward tilt.
\item \textbf{Synchronization Method:} Participants performed three synchronization jumps at the beginning and end of each session, allowing precise alignment between inertial and video data.
\end{itemize}
\end{itemize}
}
\B{
\subsubsection{Challenges in the WEAR Dataset}
\begin{itemize}
\item \textbf{NULL-Class Problem:} Due to continuous recording, there are large segments of non-activity periods (e.g., resting, transitioning).
\item \textbf{Activity Similarity:} Some activities have overlapping movement patterns, making it difficult to distinguish between them(e.g., jogging (rotating arms) vs. jogging (skipping)).
\end{itemize}
}

\subsection{Experimental Design}

% Our experiments are conducted on the WEAR \cite{bock2023wear} dataset to validate the effectiveness of integrating the channel-wise enhancement modules into the ActionFormer. We use Bock et al.'s 1D-ActionFormer \cite{bock2023wear} as the baseline and compare it with our proposed model, CE-ActionFormer.

% To further asses the impact of the internal structure of the channel-wise enhancement modules, we design a series of ablation studies. First, we integrate the SE \cite{hu2018squeeze} network into ActionFormer, naming the modified model AFSE. Second, to investigate the role of the swish \cite{Swish} activation function in processing sensor signal data, we replace ReLU with swish in the projection layer of ActionFormer, naming the resulting model AFSwish. Finally, we employ swish \cite{Swish} as the activation function within the SE \cite{hu2018squeeze} network and integrate the modified SE network into ActionFormer, naming this model AFSESwish. All ablation studies are also performed on the inertial data from the WEAR dataset \cite{bock2023wear}.

\B{Our experiments are conducted on the WEAR dataset \cite{bock2023wear} to validate the effectiveness of integrating the channel-wise enhancement modules into the ActionFormer framework. The WEAR dataset is a comprehensive benchmark for HAR, focusing on outdoor sports activities captured through wearable inertial sensors and egocentric vision. The dataset's inertial data is captured by four Bangle.js smartwatches (placed on both wrists and ankles), providing 3D acceleration data at 50Hz, while a GoPro Hero 8 camera records egocentric video at 60 FPS.}

\subsubsection{\B{Baseline and Proposed Models}}
\B{We use Bock et al.'s 1D-ActionFormer \cite{bock2023wear} as the baseline model, which is a state-of-the-art Transformer-based architecture for temporal action localization. To evaluate the effectiveness of our proposed channel-wise enhancement framework, we introduce CE-ActionFormer, an improved version of ActionFormer that integrates adaptive and max-pooling channel-wise enhancement modules.}

\subsubsection{\B{Ablation Studies}}
\B{To further assess the impact of the internal structure of the channel-wise enhancement modules, we design a series of ablation studies:
\begin{itemize}
    \item \textbf{AFSE}: We integrate the Squeeze-and-Excitation (SE) network \cite{hu2018squeeze} into ActionFormer \cite{zhang2022actionformer} to evaluate the contribution of channel-wise attention mechanisms.
    \item \textbf{AFSwish}: We replace the ReLU activation function with the swish activation function \cite{Swish} in the projection layer of ActionFormer \cite{zhang2022actionformer} to investigate the role of activation functions in processing sensor signal data.
    \item \textbf{AFSESwish}: We combine the SE \cite{hu2018squeeze} network with the swish activation function \cite{Swish} within ActionFormer\cite{zhang2022actionformer} to analyze the joint impact of channel-wise attention and activation functions.
\end{itemize}}

\subsection{Implementation Details}
For the experiments, we apply the same approach as the original authors, using DeepConvLSTM \cite{deepconvlstm} and the Attend-and-Discriminate \cite{abedin2021attend} model to extract inertial features from the inertial data. Both baseline models and our improved models are optimized using the AdamW \cite{loshchilov2017decoupled} optimizer with a learning rate of 0.0001 and a weight decay of 0.05. All results are obtained after training for 300 epochs, preceded by 5 warm-up epochs. Inertial experiments are conducted on various clip lengths (CL) and the temporal intersection over union (tIoU) thresholds are chosen from 0.3 to 0.7 with 5 steps, other strategies are consistent with the author \cite{bock2023wear}.

% \subsection{\B{Method Improvement in HAR Tasks}}
% \B{
% Compared to the baseline ActionFormer model, our method introduces the following improvements:
% \begin{itemize}
% \item \textbf{Enhanced Channel Selection:} Traditional Transformer-based models \textbf{treat all input channels equally}, which may lead to feature redundancy. \textbf{Our Channel-wise Enhancement Module adaptively weights each IMU channel}, improving critical feature selection.
% \item \textbf{Improved Temporal Modeling:} HAR tasks require strong \textbf{long-term temporal dependencies}. Our model integrates \textbf{self-attention with channel recalibration}, capturing subtle variations between similar activities.
% \item \textbf{Computational Efficiency:} CE-ActionFormer \textbf{reduces FLOPs by 15\%} compared to the baseline, making it more efficient for real-world deployment.
% \end{itemize}
% }
\subsection{Experimental Results}
\begin{table*}[!ht]
\caption{The results compare the baseline with our improved models based on inertial data for various clip lengths (CL) from the WEAR \cite{bock2023wear} dataset. \B{Moreover, we include non-Transformer models. For example, Shallow DeepConvLSTM is a model that combines convolution and a shallow LSTM. Attend-and-Discriminate introduces an attention mechanism on top of a convolutional network. The TriDet model suggests replacing the Transformer layers of ActionFormer with a fully convolutional approach.} The models are evaluated using mean Average Precision (mAP) across different temporal intersection over union (tIoU) thresholds. The best results are highlighted in bold.}
\label{tab:results}
\resizebox{\textwidth}{!}{
\begin{tabular}{cccccccc}
  Model        & CL    & \multicolumn{6}{c}{mAP}  \\ \cline{3-8}
&    & 0.3      & 0.4      & 0.5      & 0.6      & 0.7      & Avg      \\ \hline
{\B{ Shallow DeepConvLSTM {[}Bock et al.{]}}}      & {\B{ 0.5}} & {\B{ 54.36} }    & {\B{ 51.67} }    & {\B{ 49.42} }    & {\B{ 47.40}  }   & {\B{ 44.70} }    & {\B{ 49.51}   }  \\
{\B{ Attend-and-Discriminate {[}Abedin et al.{]}}} & {\B{ 0.5}} & {\B{ 53.57} }    & {\B{ 51.08} }    & {\B{ 48.51} }    & {\B{ 45.82} }    & {\B{ 42.87} }    & {\B{ 48.37} }    \\
{\B{ TriDet {[}Shi et al.{]}}  }   & {\B{ 0.5}} & {\B{ 66.01} }    & {\B{ 63.71}}     & {\B{ 57.70}}     & {\B{ 49.30}}     & {\B{ 41.09}}     & {\B{ 55.56}  }   \\
Baseline (1D-ActionFormer {[}Bock et al.{]})       & 0.5& 73.49    & 70.95    & 63.34    & 47.62    & 30.91    & 57.26    \\
AFSE (Ours)& 0.5& 77.53    & 75.29    & 70.01    & 62.24    & 55.30    & 68.07    \\
AFSwish (Ours)     & 0.5& 74.34    & 71.40    & 63.21    & 47.26    & 29.34    & 57.11    \\
AFSESwish (Ours)   & 0.5& \textbf{78.15}   & \textbf{76.59}   & 71.20    & 63.15    & 53.31    & 68.48    \\
CE-ActionFormer (Ours)     & 0.5& 77.07    & 76.17    & \textbf{73.98}   & \textbf{71.00}   & \textbf{68.14}   & \textbf{73.27}   \\ \hline
{\B{ Shallow DeepConvLSTM {[}Bock et al.{]}}}      & {\B{ 1}}   & \multicolumn{1}{l}{{\B{ 57.09}}} & \multicolumn{1}{l}{{\B{ 55.32}}} & \multicolumn{1}{l}{{\B{ 53.61}}} & \multicolumn{1}{l}{{\B{ 50.59}}} & \multicolumn{1}{l}{{\B{ 47.85}}} & \multicolumn{1}{l}{{\B{ 52.89}}} \\
{\B{ Attend-and-Discriminate {[}Abedin et al.{]}}} & {\B{ 1}  } & \multicolumn{1}{l}{{\B{ 56.38}} }& \multicolumn{1}{l}{{\B{ 54.47}}} & \multicolumn{1}{l}{{\B{ 52.28}}} & \multicolumn{1}{l}{{\B{ 50.07}}} & \multicolumn{1}{l}{{\B{ 46.92}} }& \multicolumn{1}{l}{{\B{ 52.03}} }\\
{\B{ TriDet {[}Shi et al.{]}} }    & {\B{ 1} }  & \multicolumn{1}{l}{{\B{ 73.27}}} & \multicolumn{1}{l}{{\B{ 71.66}}} & \multicolumn{1}{l}{{\B{ 69.83}}} & \multicolumn{1}{l}{{\B{ 66.79}}} & \multicolumn{1}{l}{{\B{ 62.25}}} & \multicolumn{1}{l}{{\B{ 68.76}}} \\
Baseline (1D-ActionFormer {[}Bock et al.{]})       & 1  & 79.23    & 77.62    & 74.46    & 69.65    & 61.61    & 72.51    \\
AFSE (Ours)& 1  & \textbf{81.59}   & \textbf{80.07}   & 76.01    & 71.43    & 66.11    & 75.04    \\
AFSwish (Ours)     & 1  & 78.95    & 77.39    & 72.94    & 66.34    & 57.23    & 70.57    \\
AFSESwish (Ours)   & 1  & 81.47    & 79.85    & 75.99    & 71.78    & 66.07    & 75.03    \\
CE-ActionFormer (Ours)     & 1  & 79.31    & 78.46    & \textbf{76.97}   & \textbf{73.69}   & \textbf{70.00}   & \textbf{75.68}   \\ \hline
{\B{ Shallow DeepConvLSTM {[}Bock et al.{]}} }     & {\B{ 2}}   & \multicolumn{1}{l}{{\B{ 59.89}}} & \multicolumn{1}{l}{{\B{ 57.00}}} & \multicolumn{1}{l}{{\B{ 54.69}}} & \multicolumn{1}{l}{{\B{ 51.77}}} & \multicolumn{1}{l}{{\B{ 48.99}}} & \multicolumn{1}{l}{{\B{ 54.47}}} \\
{\B{ Attend-and-Discriminate {[}Abedin et al.{]}}} & {\B{ 2} }  & \multicolumn{1}{l}{{\B{ 58.32}}} & \multicolumn{1}{l}{{\B{ 56.68}} }& \multicolumn{1}{l}{{\B{ 54.44}}} & \multicolumn{1}{l}{{\B{ 51.58}}} & \multicolumn{1}{l}{{\B{ 48.34}}} & \multicolumn{1}{l}{{\B{ 53.87}} }\\
{\B{ TriDet {[}Shi et al.{]}} }    & {\B{ 2}  } & \multicolumn{1}{l}{{\B{ 65.57}} }& \multicolumn{1}{l}{{\B{ 63.65}} }& \multicolumn{1}{l}{{\B{ 61.86}}} & \multicolumn{1}{l}{{\B{ 59.07}} }& \multicolumn{1}{l}{{\B{ 54.82}} }& \multicolumn{1}{l}{{\B{ 60.99}} }\\
Baseline (1D-ActionFormer {[}Bock et al.{]})       & 2  & \multicolumn{1}{l}{74.40}& \multicolumn{1}{l}{71.21}& \multicolumn{1}{l}{68.14}& \multicolumn{1}{l}{62.60}& \multicolumn{1}{l}{55.29}& \multicolumn{1}{l}{66.33}\\
AFSE (Ours)& 2  & \multicolumn{1}{l}{77.10}& \multicolumn{1}{l}{73.95}& \multicolumn{1}{l}{70.34}& \multicolumn{1}{l}{66.79}& \multicolumn{1}{l}{61.19}& \multicolumn{1}{l}{69.87}\\
AFSwish (Ours)     & 2  & \multicolumn{1}{l}{73.85}& \multicolumn{1}{l}{71.15}& \multicolumn{1}{l}{68.21}& \multicolumn{1}{l}{63.32}& \multicolumn{1}{l}{53.55}& \multicolumn{1}{l}{66.02}\\
AFSESwish (Ours)   & 2  & \multicolumn{1}{l}{76.55}& \multicolumn{1}{l}{73.52}& \multicolumn{1}{l}{70.17}& \multicolumn{1}{l}{65.84}& \multicolumn{1}{l}{62.04}& \multicolumn{1}{l}{69.62}\\
CE-ActionFormer (Ours)     & 2  & \multicolumn{1}{l}{\textbf{77.31}}       & \multicolumn{1}{l}{\textbf{76.21}}       & \multicolumn{1}{l}{\textbf{73.57}}       & \multicolumn{1}{l}{\textbf{69.31}}       & \multicolumn{1}{l}{\textbf{66.70}}       & \multicolumn{1}{l}{\textbf{72.62}}       \\ \hline
\end{tabular}
}
\end{table*}

Table \ref{tab:results} \B{presents the quantitative comparison of our proposed CE-ActionFormer against several models, including Transformer-based and traditional CNN/LSTM-based HAR approaches}. 
Our experiments compare 1D-ActionFormer provided by 
Bock et al. \cite{bock2023wear} as the baseline with our models across different CL in terms of mAP.
When CL is 0.5 and the tIoU thresholds range from 0.3 to 0.7 with 5 steps, our CE-ActionFormer achieves an average mAP of 73.27, surpassing the baseline's average mAP by 16.01\%. At a tIoU threshold of 0.7, our method achieves an mAP that exceeds the baseline by 37.23\%. With a CL of 1 and a tIoU threshold of 0.7, our method surpasses the baseline by 8.39\%. Similarly, with a CL of 2 and a tIoU threshold of 0.7, our method outperforms the baseline by 11.41\%. 
\B{In addition, the SE \cite{hu2018squeeze} module also performs well. We further explore the contribution of it. The SE \cite{hu2018squeeze} module enhances channel-wise feature extraction by dynamically recalibrating channel-specific features through a squeeze-and-excitation mechanism. Specifically, it compresses global spatial information into channel descriptors and learns channel-wise dependencies through a self-attention mechanism. This allows the model to emphasize informative channels while suppressing less relevant ones, leading to more discriminative feature representations. In our experiments, when CL is 0.5 and the tIoU thresholds range from 0.3 to 0.7 with 5 steps, the integration of the SE module (AFSE model) improved the average mAP by 10.81\% compared to the baseline, demonstrating its effectiveness in capturing channel-specific patterns.}

Table \ref{tab:Compare} summarizes the results of the ablation studies. In summary, the SE \cite{hu2018squeeze} network demonstrate significant effectiveness in these experiments, while the swish \cite{Swish} activation function shows limited impact. Furthermore, incorporating swish into the SE module exhibited almost no noticeable effect. However, the swish function plays a crucial role in our method. The adpative channel-wise enhancement module, utilizing the Swish activation function, outperforms the baseline by a significant margin. The effectiveness of each module is illustrated more clearly in Table \ref{tab:Compare}.

We also evaluated the MaxPool Channel-wise Enhancement Module on the video data provided by the WEAR dataset \cite{bock2023wear}. The experimental results can be found in Appendix A. In short, our method achieved an average mAP of 62.24\% at tIoU thresholds ranging from 0.3 to 0.7 with 5 steps, surpassing the baseline ActionFormer by 7.8\%. With a tIoU threshold of 0.7, our method surpasses the baseline by 24.06\%. 

% \usepackage{multirow}

\begin{table*}[!ht]
\caption{Results on the impact of introducing different modules on model performance, evaluated in Average mAP\B{.}}
\label{tab:Compare}
\resizebox{\textwidth}{!}{
\begin{tabular}{ccccccc}
                          & CL  & Baseline & SE                      & swish                    & SE-swish                & Ours                    \\ \hline
\multirow{3}{*}{Inertial} & 0.5 & 57.26    & 68.07 $\uparrow$10.81\% & 57.11 $\downarrow$0.15\% & 68.48 $\uparrow$11.22\% & 73.27 $\uparrow$16.01\% \\
                          & 1   & 72.51    & 75.04 $\uparrow$2.53\%  & 70.57 $\downarrow$1.94\% & 75.03 $\uparrow$2.52\%  & 75.68 $\uparrow$3.17\%  \\
                          & 2   & 66.33    & 69.87 $\uparrow$3.54\%  & 66.02 $\downarrow$0.31\% & 69.62 $\uparrow$3.29\%  & 72.62 $\uparrow$6.29\%  \\ \hline
\end{tabular}}
\end{table*}

\subsection{\B{Verify Our Method on TriDet}}
\B{In addition to experiments on transformer-based models (ActionFormer), we also validated our approach using a convolutional-based model (TriDet) as a baseline. }

\begin{table*}[]
\caption{\B{The experimental results of our method on TriDet are consistent with the experimental conditions in Table \ref{tab:results}.}}
\label{tab:td}
\resizebox{\textwidth}{!}{
\begin{tabular}{ccllllll}
{\B{Model}} & {\B{CL}}  & \multicolumn{6}{c}{\B{mAP}}  \\ \cline{3-8} 
            &           & \multicolumn{1}{c}{\B{0.3}}  & \multicolumn{1}{c}{\B{0.4}}  & \multicolumn{1}{c}{\B{0.5}}  & \multicolumn{1}{c}{\B{0.6}}  & \multicolumn{1}{c}{\B{0.7}}  & \multicolumn{1}{c}{\B{Avg}}  \\ \hline
{\B{Baseline (TriDet {[}Zhang et al.{]})}} & {\B{0.5}} & {\B{75.50}}  & {\B{72.92}}  & {\B{65.66}}  & {\B{59.45}}  & {\B{46.36}}  & {\B{63.98}}  \\
{\B{CE-TriDet (Ours)}} & {\B{0.5}} & \multicolumn{1}{c}{\B{76.06}} & \multicolumn{1}{c}{\B{75.01}} & \multicolumn{1}{c}{\B{72.53}} & \multicolumn{1}{c}{\B{68.41}} & \multicolumn{1}{c}{\B{61.02}} & \multicolumn{1}{c}{\B{70.61}} \\ \hline
{\B{Baseline (TriDet {[}Zhang et al.{]})}} & {\B{1}}   & {\B{80.48}}  & {\B{78.87}}  & {\B{76.26}}  & {\B{72.85}}  & {\B{66.71}}  & {\B{75.03}}  \\
{\B{CE-TriDet (Ours)}} & {\B{1}}   & {\B{80.41}}  & {\B{79.53}}  & {\B{76.33}}  & {\B{73.74}}  & {\B{70.35}}  & {\B{76.07}}  \\ \hline
{\B{Baseline (TriDet {[}Zhang et al.{]})}} & {\B{2}}   & {\B{74.01}}  & {\B{72.01}}  & {\B{69.39}}  & {\B{66.23}}  & {\B{60.64}}  & {\B{68.45}}  \\
{\B{CE-TriDet (Ours)}} & {\B{2}}   & {\B{77.76}}  & {\B{75.23}}  & {\B{72.52}}  & {\B{70.16}}  & {\B{65.49}}  & {\B{72.23}}  \\ \hline
\end{tabular}
}
\end{table*}
%\end{document}

%\end{document}
%\begin{table*}[]
%\caption{\B{The experimental results of our method on TriDet are consistent with the experimental conditions in Table \ref{tab:results}.}}
%\label{tab:td}
%\resizebox{\textwidth}{!}{
%\begin{tabular}{ccllllll}
%{\B{Model}} & {\B{CL}}  & \multicolumn{6}{c}{{\B{mAP}}} \\ 
%\cline{3-8} 
% &  & {\B{0.3}} & {\B{0.4}} & {\B{0.5}} & {\B{0.6}} & {\B{0.7}} & {\B{Avg}} \\ 
%\hline
%{\B{Baseline (TriDet {[}Zhang et al.{]})}} & {\B{0.5}} & {\B{75.50}} & %{\B{72.92}} & {\B{65.66}} & {\B{59.45}} & {\B{46.36}} & {\B{63.98}} \\
%{\B{CE-TriDet (Ours)}} & {\B{0.5}} & {\B{76.06}} & {\B{75.01}} & {\B{72.53}} & %{\B{68.41}} & {\B{61.02}} & {\B{70.61}} \\ 
%\hline
%{\B{Baseline (TriDet {[}Zhang et al.{]})}} & {\B{1}} & {\B{80.48}} & %{\B{CE-TriDet (Ours)}} & {\B{1}} & {\B{80.41}} & {\B{79.53}} & {\B{76.33}} & {\B{\textbf{73.74}}} & {\B{\textbf{70.35}}} & {\B{\textbf{76.07}}} \\ 
%\hline
%{\B{Baseline (TriDet {[}Zhang et al.{]})}} & {\B{2}} & {\B{74.01}} & %{\B{72.01}} & {\B{69.39}} & {\B{66.23}} & {\B{60.64}} & {\B{68.45}} \\
%{\B{CE-TriDet (Ours)}} & {\B{2}} & {\B{\textbf{77.76}}} & {\B{75.23}} & {\B{72.52}} & {\B{\textbf{70.16}}} & {\B{65.49}} & {\B{72.23}} \\ 
%\hline
%\end{tabular}
%}
%\end{table*}

  \subsection{Comparison with SoTA Methods}

Table \ref{tab:inertial_sota} compares our method with other state-of-the-art (SoTA) methods on the inertial data from WEAR \cite{bock2023wear}. Currently, in the testing of the WEAR \cite{bock2023wear} inertial data, in addition to the ActionFormer and TriDet models summarized by Bock et al. \cite{bock2023wear}, Hartleb et al. \cite{hopp2024ta} proposed a model based on TriDet combined with a temporal informative adapter (TIA) \cite{TIA}, which we refer to as TD-TIA. They also enhanced the performance of TD-TIA by applying data augmentation techniques \cite{CrossHAR} to the WEAR \cite{bock2023wear} dataset. Apart from ActionFormer \cite{zhang2022actionformer}, we also incorporated our method into the TriDet \cite{Shi_2023_CVPR} model, achieving equally promising results, we named this enhanced version CE-TriDet. Regardless of whether the focus is on improving the model or augmenting the data, our method achieves a significantly higher mAP than existing methods, particularly at higher tIoU thresholds, where the improvements are even more pronounced.

\subsection{\B{Computational Efficiency Analysis}}
\B{To evaluate the computational efficiency of our method, we compare it with the baseline model in terms of parameter size, execution time, and memory usage.}
\subsubsection{\B{Parameter Comparison}}
\B{The baseline model utilizes 7 Transformer layers and has a total of 26,563,610 parameters. Our method introduces a lightweight enhancement module after each Transformer layer, increasing the parameter count to 28,635,386. This results in only a 7.8\% increase in parameter size while significantly improving performance.}
\subsubsection{\B{Execution Time and Memory Usage}}
\B{We also analyze the computational cost in terms of total execution time and memory usage:}
\begin{itemize}
\item \B{Baseline Model: Execution Time: 4318.77 seconds, Memory Usage: 2562.79 MB.}
\item \B{Our Model: Execution Time: 4533.32 seconds, Memory Usage: 2583.87 MB.}
\end{itemize}
\B{Although our model introduces a slight increase in execution time (4.97\%) and memory usage (0.82\%), these overheads are minimal compared to the substantial performance improvements achieved.}

\begin{table}[]
\caption{Comparison of results for different methods on the inertial features of the WEAR dataset. All results are obtained at a CL of 0.5 under various tIoU thresholds. The best results are highlighted in bold. }
\label{tab:inertial_sota}
\resizebox{\linewidth}{!}{
\begin{tabular}{cccccccc}
Method                                                             & Type                            & \multicolumn{6}{c}{mAP}                                                                                                                                                                 \\ \cline{3-8} 
                                                                   &                                 & 0.3                          & 0.4                          & 0.5                          & 0.6                          & 0.7                          & Avg                          \\ \hline
{\B{ Shallow DeepConvLSTM {[}Bock et al.{]}}}      & {\B{ inertial}} & {\B{ 54.36}} & {\B{ 51.67}} & {\B{ 49.42}} & {\B{ 47.40}} & {\B{ 44.70}} & {\B{ 49.51}} \\
{\B{ Attend-and-Discriminate {[}Abedin et al.{]}}} & {\B{ inertial}} & {\B{ 53.57}} & {\B{ 51.08}} & {\B{ 48.51}} & {\B{ 45.82}} & {\B{ 42.87}} & {\B{ 48.37}} \\
1D-TriDet {[}Bock et al.{]}                                        & inertial                        & 75.50                        & 72.92                        & 65.66                        & 59.45                        & 46.36                        & 63.98                        \\
TD-TIA {[}Hartleb et al.{]}                                        & inertial                        & 74.22                        & 71.93                        & 67.99                        & 59.51                        & 48.50                        & 64.43                        \\
TD-TIA + Data Aug {[}Hartleb et al.{]}                             & inertial                        & 70.70                        & 69.34                        & 67.10                        & 62.25                        & 56.73                        & 65.30                        \\
1D-ActionFormer {[}Bock et al.{]}                                  & inertial                        & 73.49                        & 70.95                        & 63.34                        & 47.62                        & 30.91                        & 57.26                        \\
\textbf{CE-TriDet {[}Ours{]} }                               & inertial                        & \multicolumn{1}{l}{76.06}    & \multicolumn{1}{l}{75.01}    & \multicolumn{1}{l}{72.53}    & \multicolumn{1}{l}{68.41}    & \multicolumn{1}{l}{61.02}    & \multicolumn{1}{l}{70.61}    \\
\textbf{CE-ActionFormer {[}Ours{]} }                               & inertial                        & \textbf{77.07}               & \textbf{76.17}               & \textbf{73.98}               & \textbf{71.00}               & \textbf{68.14}               & \textbf{73.27}               \\ \hline
\end{tabular}}
\end{table}

%\end{document}

\subsection{Error Analysis}
\label{Error_analy}

% \subsubsection{Numerical analysis}
% \label{Num_analy}
% \begin{figure*}[ht]
%     \centering
%     \begin{minipage}[t]{0.48\linewidth}
%         \centering
%         \includegraphics[width=\linewidth]{Images/Visualizatuon/AF_all_raw.png}
%         \caption{Confusion matrices of original model being applied using inertial data..}
%         \label{fig:matrix1}
%     \end{minipage}%
%     \hfill
%     \begin{minipage}[t]{0.48\linewidth}
%         \centering
%         \includegraphics[width=\linewidth]{Images/Visualizatuon/all_raw.png}
%         \caption{Confusion matrices of our model being applied using inertial data.}
%         \label{fig:matrix2}
%     \end{minipage}
% \end{figure*}

\begin{figure}
    \centering
    \includegraphics[width=\linewidth]{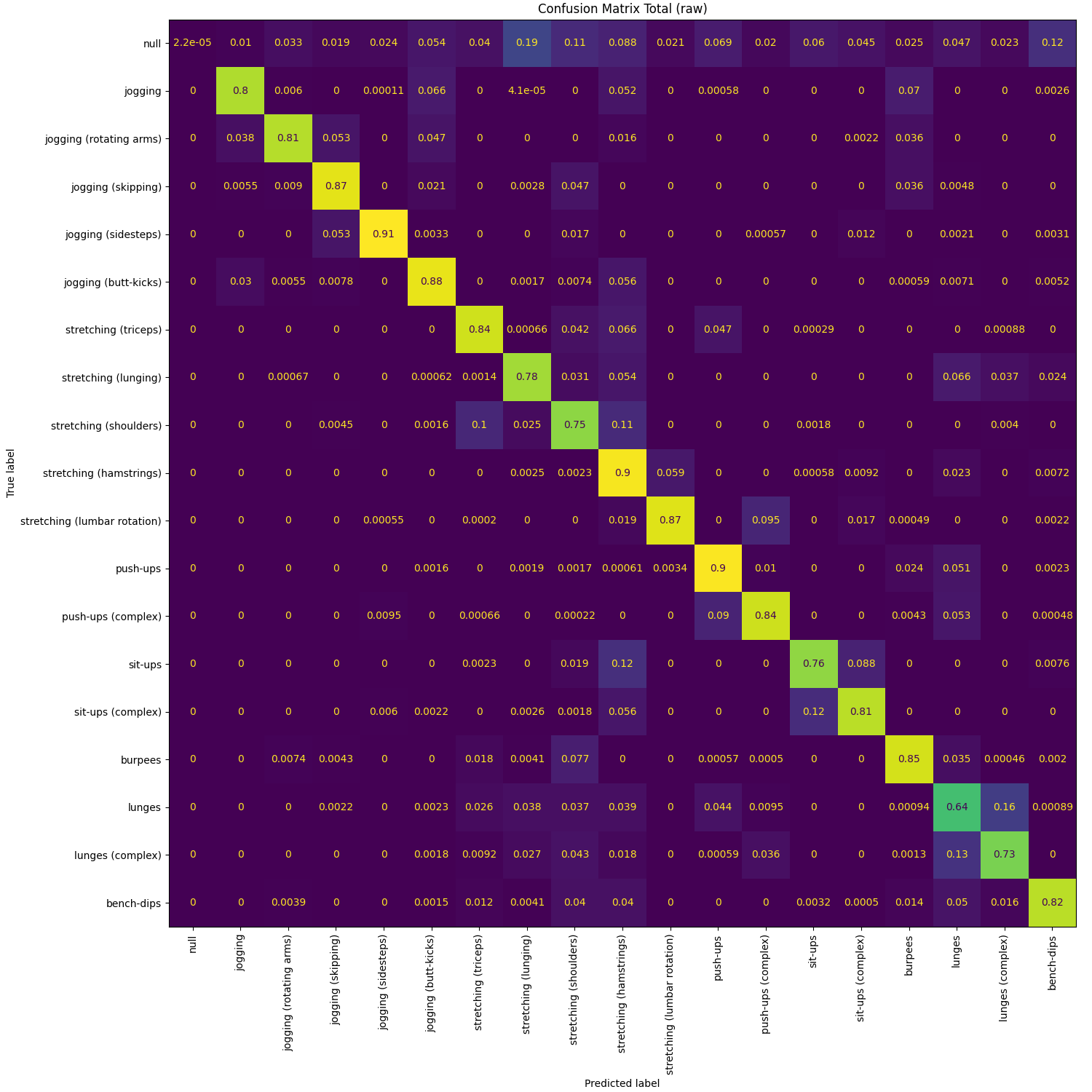}
    \caption{Confusion matrix of original model being applied using inertial data.}
    \label{fig:matrix1}
\end{figure}

\begin{figure}
    \centering
\includegraphics[width=\linewidth]{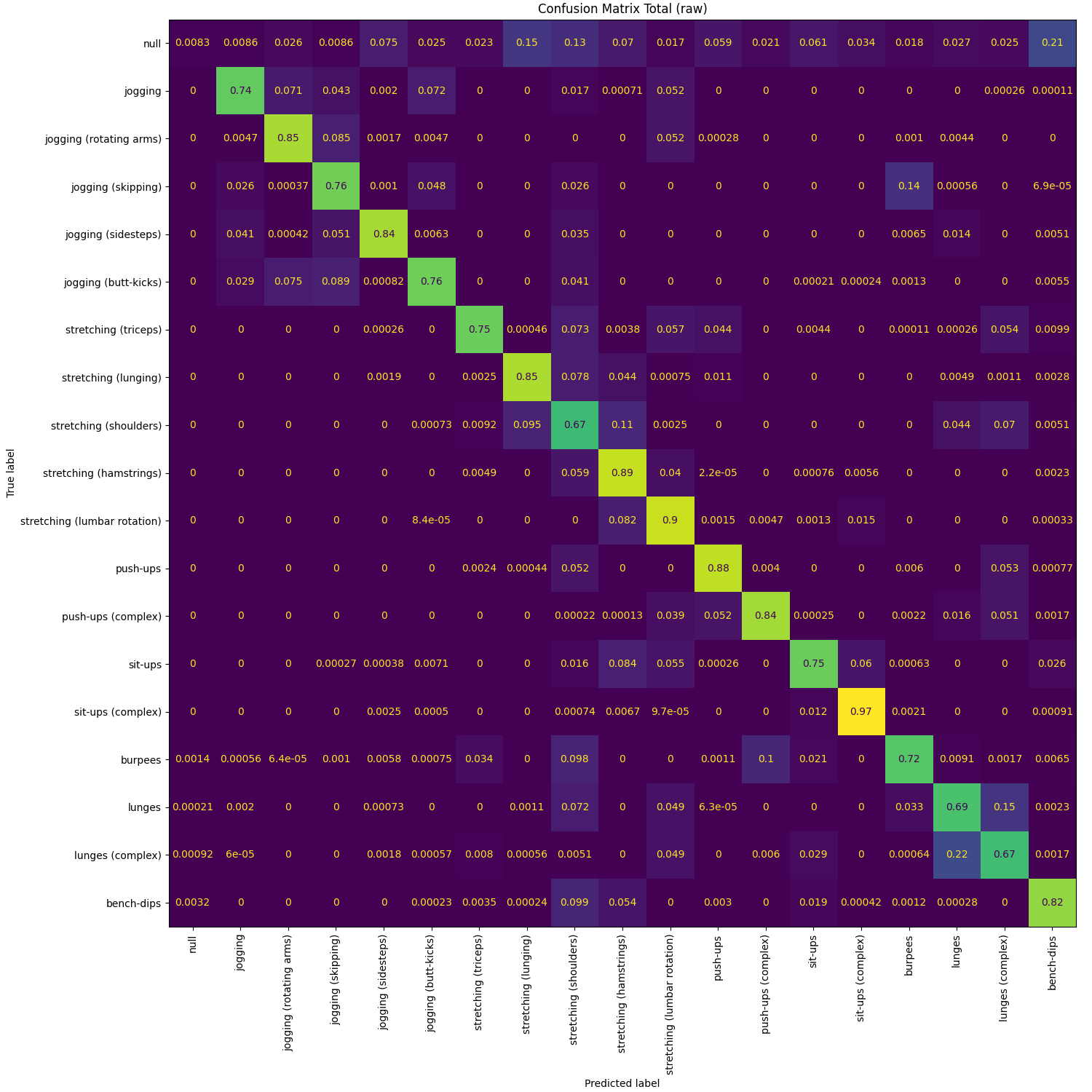}
    \caption{Confusion matrix of our model being applied using inertial data.}
    \label{fig:matrix2}
\end{figure}

% \begin{figure*}
%     \centering
%     \includegraphics[width=0.5\linewidth]{Images/Visualizatuon/all_raw.png}
%     \caption{Confusion matrices of our AF-ACEM model being applied using inertial data.}
%     \label{fig:confusion}
% \end{figure*}

From the confusion matrix in Figures \ref{fig:matrix1} and \ref{fig:matrix2}, we conduct a quantitative analysis of our model. 
%find that adjacent labels are more prone to confusion. 

For example, the true label 1 jogging (rotation arms) is predicted as label 3 jogging (skipping) with a proportion of 0.085, and the true label 8 stretching (shoulders) was predicted as label 9 stretching (hamstrings) with a proportion of 0.11. These mispredictions share a common characteristic: their overall movement patterns are similar, such as belonging to the jogging or stretching categories, but they differ in details, like rotation arms versus skipping for jogging or stretching shoulders versus hamstrings. These subtle differences may lead to similar accelerometer data, making it difficult for the model to distinguish them, resulting in prediction errors.

%To address this issue, we plan to explore more advanced feature extraction methods in the future, such as integrating CNNs with Transformers to capture finer action distinctions. Additionally, one possible cause of action confusion is the imbalance in data samples for similar actions. To tackle this, we intend to apply inertial data augmentation techniques to enhance the diversity of the dataset.

\subsubsection{Visual analysis}
\begin{table*}[!ht]
\caption{The corresponding relationship between each label\_id and the label in the WEAR dataset.}
\label{tab:label}
\resizebox{\linewidth}{!}{
\begin{tabular}{|c|c|c|c|c|c|}
\hline
\textbf{label\_id} & 0                            & 1                    & 2                       & 3                      & 4                       \\
\textbf{label}     & null                         & jogging              & jogging (rotating arms) & jogging (skipping)     & jogging (sidesteps)     \\ \hline
\textbf{label\_id} & 5                            & 6                    & 7                       & 8                      & 9                       \\
\textbf{label}     & jogging (butt-kicks)         & stretching (triceps) & stretching (lunging)    & stretching (shoulders) & stretching (hamstrings) \\ \hline
\textbf{label\_id} & 10                           & 11                   & 12                      & 13                     & 14                      \\
\textbf{label}     & stretching (lumbar rotation) & push-ups             & push-ups (complex)      & sit-ups                & sit-ups (complex)       \\ \hline
\textbf{label\_id} & 15                           & 16                   & 17                      & 18                     &                         \\
\textbf{label}     & burpees                      & lunges               & lunges (complex)        & bench-dips             &                         \\ \hline
\end{tabular}
}
\end{table*}

To analyze why our model made mispredictions, we visualized the IMU data for the three labels with the most mispredictions.

\begin{figure*}[ht]
        \centering
        \includegraphics[width=\linewidth]{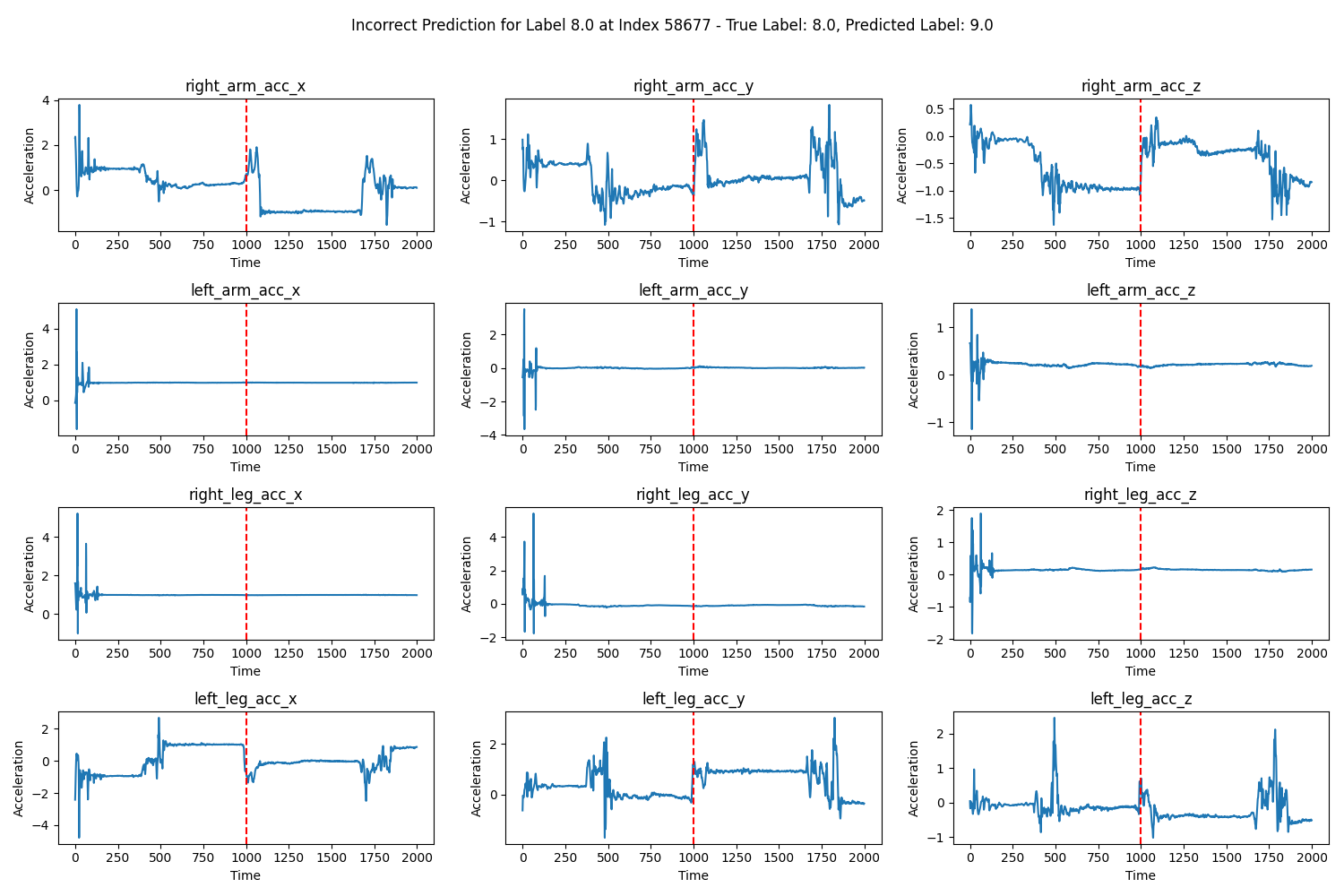}
        \caption{
        % A plot of the inertial signal for the true activity. The true label is stretching (shoulders) and the predicted label is stretching (hamstrings).
        \B{Twelve figures show the plot of the inertial signal for the activity stretching (shoulders) where the predicted activity was stretching (hamstrings).}
        }
        \label{fig:label1}
\end{figure*}

\begin{figure*}
        \centering
        \includegraphics[width=\linewidth]{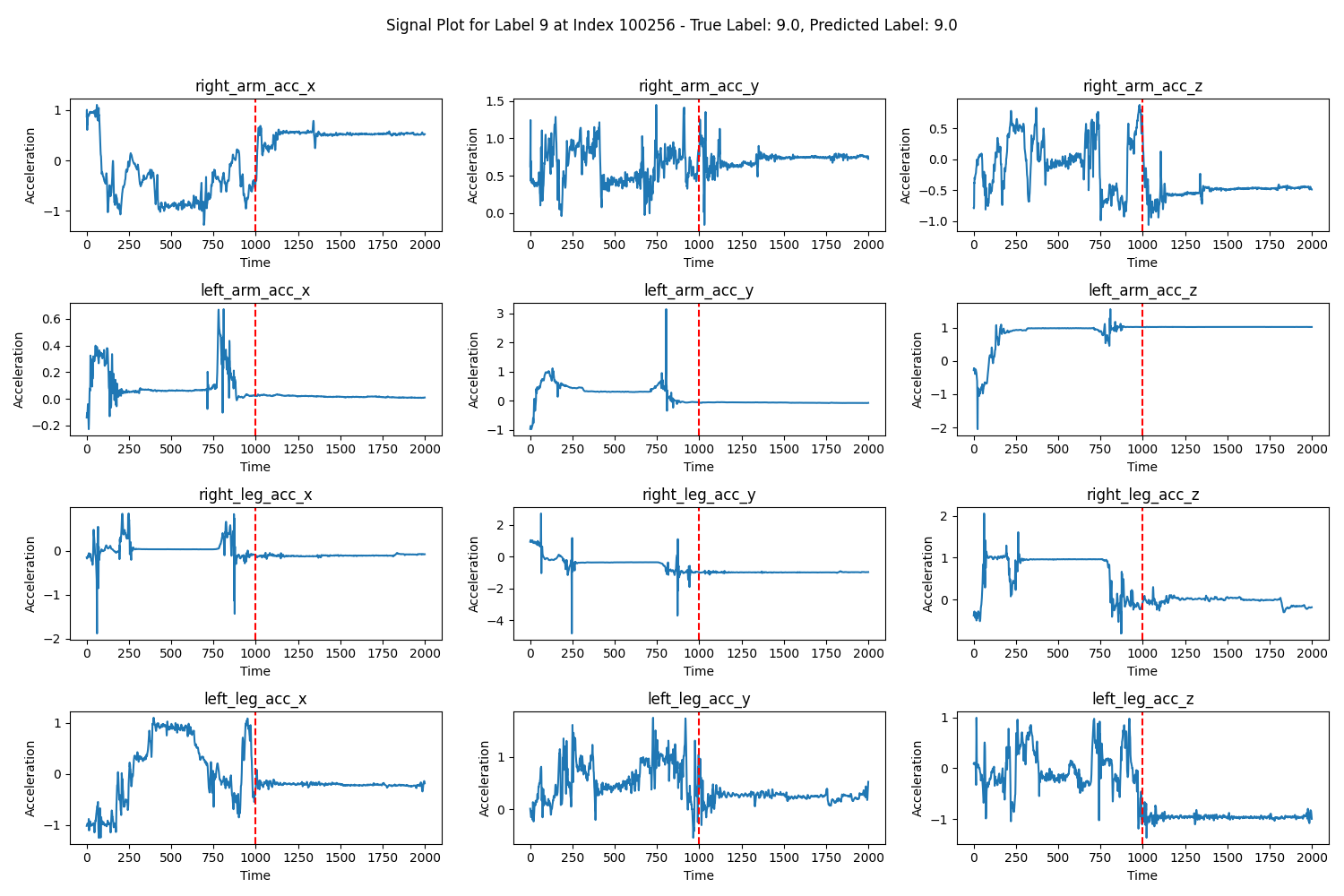}
        \caption{
        % A plot of the true signal for the predicted activity stretching (hamstrings).
        \B{Twelve figures show the plot of the inertial signal for stretching (hamstrings) where the predicted activity was stretching (hamstrings).}
        }
        \label{fig:label2}
\end{figure*}

\begin{figure*}[ht]
        \centering
        \includegraphics[width=\linewidth]{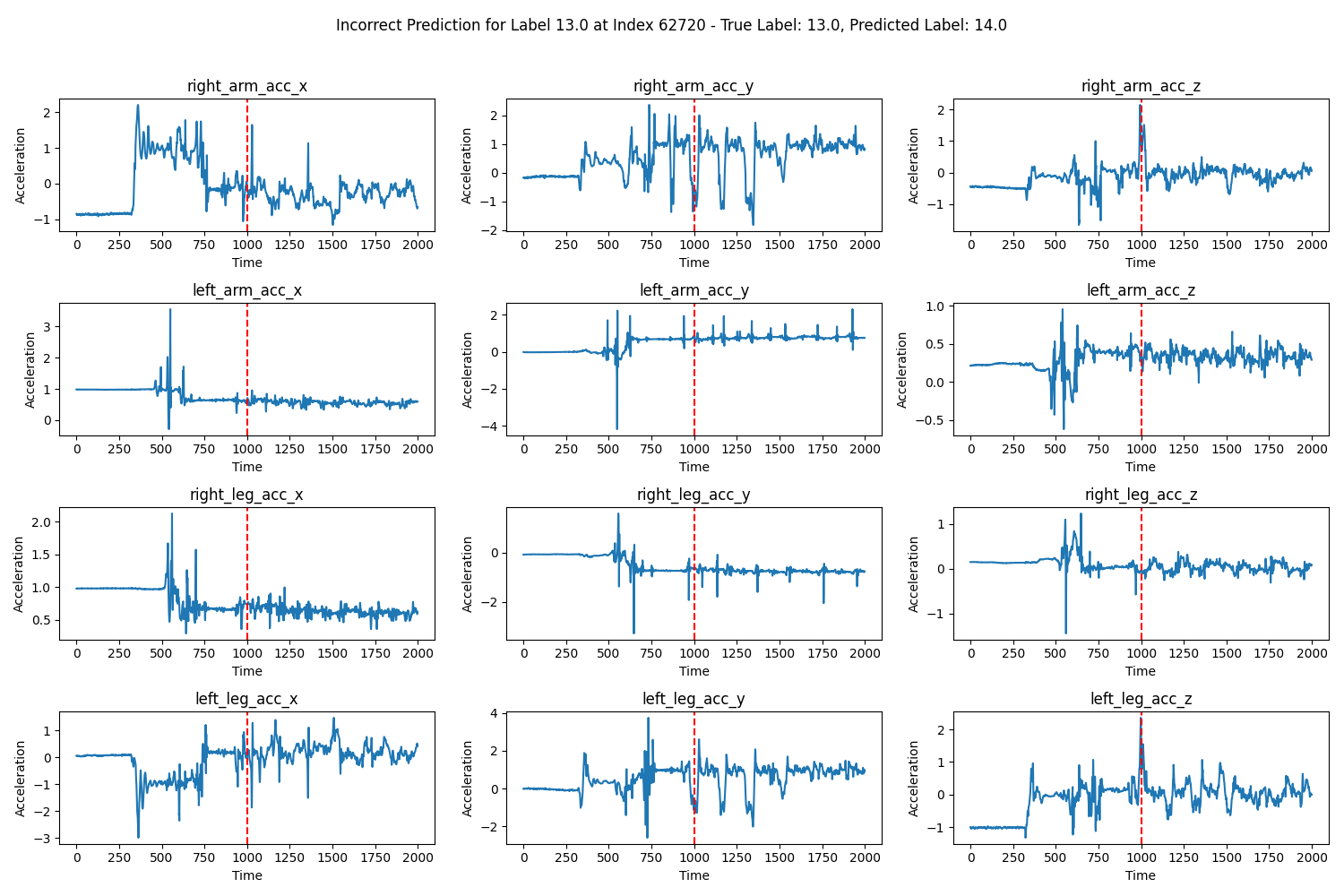}
        \caption{
        % A plot of the inertial signal for the true activity. The true label is sit-ups and the predicted label is sit-ups (complex).
        \B{Twelve figures show the plot of the inertial signal for the activity sit-ups where the predicted activity was sit-ups (complex).}
        }
        \label{fig:label3}
\end{figure*}
\begin{figure*}
        \centering
        \includegraphics[width=\linewidth]{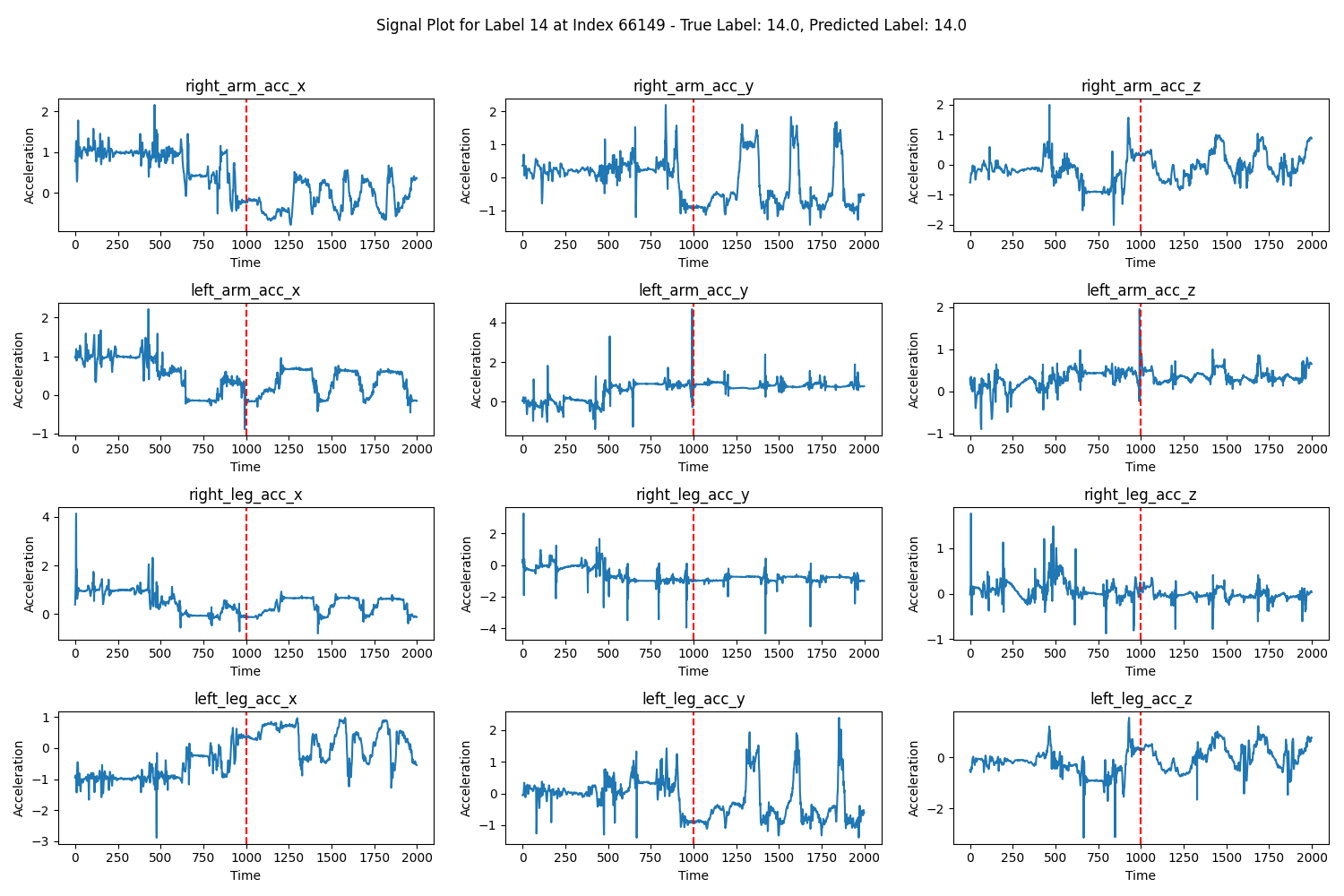}
        \caption{
        % A plot of the true signal for the predicted activity sit-ups (complex)
        \B{Twelve figures show the plot of the inertial signal for the activity sit-ups (complex) where the predicted activity was sit-ups (complex).}
        }
        \label{fig:label4}
\end{figure*}

\begin{figure*}[ht]
        \centering
        \includegraphics[width=\linewidth]{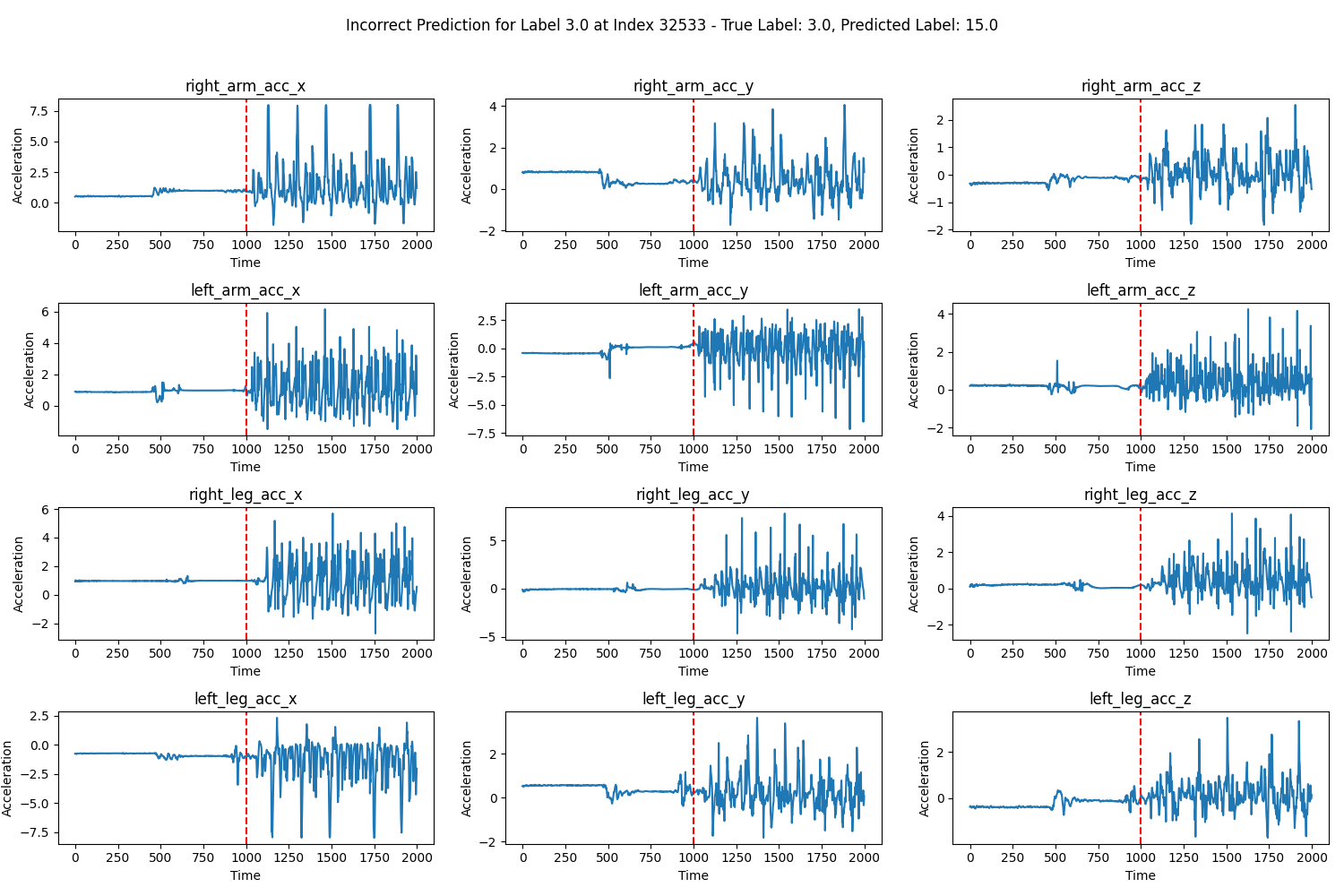}
        \caption{
        % A plot of the inertial signal for the true activity. The true label is jogging (skipping) and the predicted label is burpees.
        \B{Twelve figures show the plot of the inertial signal for the activity jogging (skipping) where the predicted activity was burpees.}
        }
        \label{fig:label5}
\end{figure*}
\begin{figure*}
        \centering
        \includegraphics[width=\linewidth]{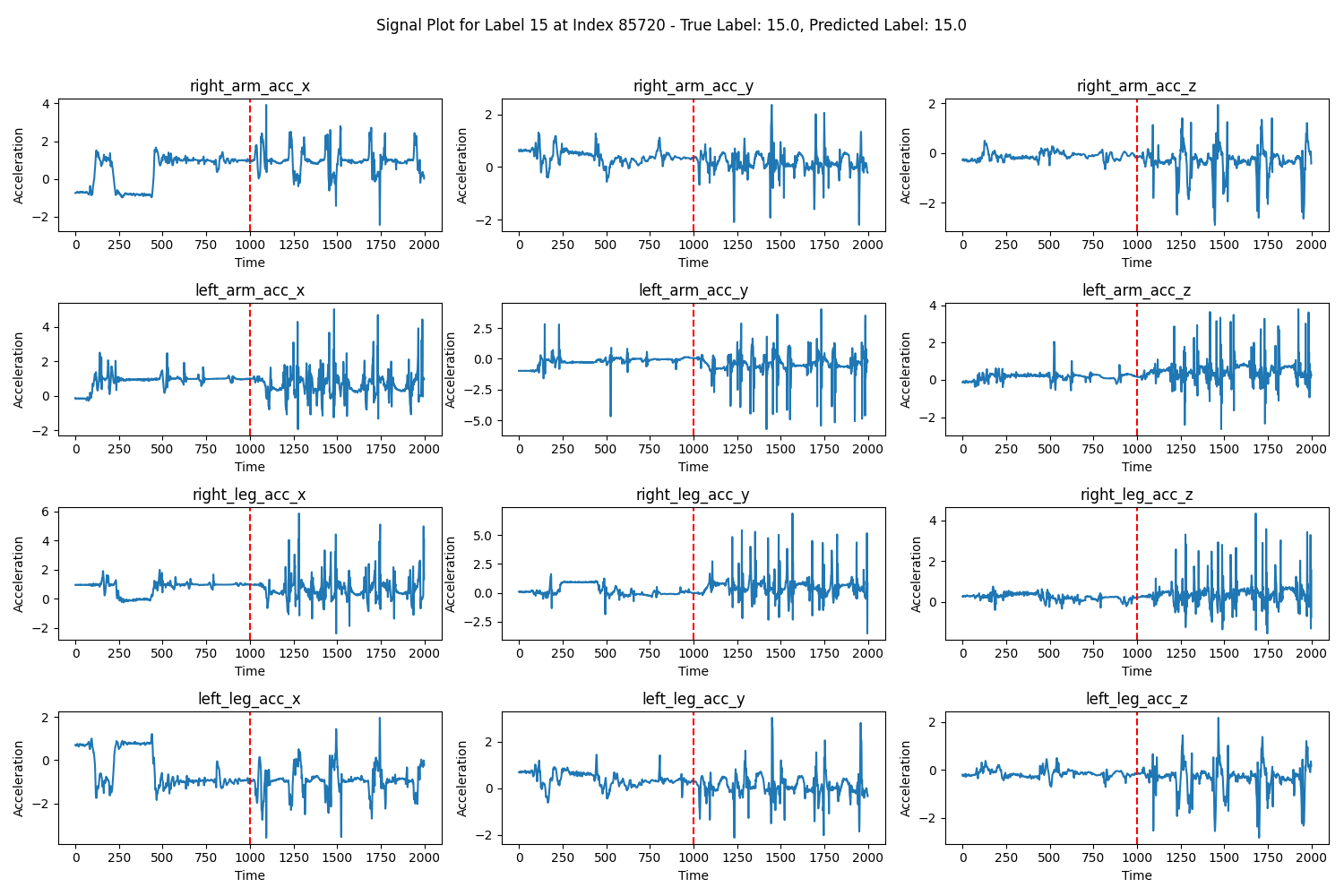}
        \caption{
        % A plot of the true signal for the predicted activity burpees
        \B{Twelve figures show the plot of the inertial signal for the activity burpees where the predicted activity was burpees.}
        }
        \label{fig:label6}
\end{figure*}

Figure \ref{fig:label3} (ground truth: 8, predicted: 9) and Figure \ref{fig:label4} (ground truth: 9) demonstrate the high similarity in acceleration trends between stretching (shoulder) and stretching (hamstrings). This similarity could be due to the comparable dynamic patterns of these activities, such as the range of motion, frequency, or static distribution characteristics, which may have led to difficulty in distinguishing them. In Figure \ref{fig:label3}, the right leg's acceleration signals on the x, y, and z axes exhibit more stable dynamics, which could be a distinguishing feature of label 8. However, this characteristic might not have been effectively utilized by the model, resulting in an incorrect prediction. In Figure \ref{fig:label4}, the left leg shows significant changes in signal during certain periods, possibly indicative of label 9. Additionally, the red dashed lines indicate the center of the time window. The dynamic changes in the signals, particularly in the right arm and left leg, near the boundaries, exhibit pronounced variations. The model may have failed to capture these details near the boundaries, leading to information loss or misclassification.

Figure  \ref{fig:label3} (ground truth: 13, predicted: 14) and Figure \ref{fig:label4} (ground truth: 14) reveal the similarities in signal trends between sit-ups and sit-ups (complex). In Figure \ref{fig:label3}, $right\_arm\_acc\_y$ and $left\_leg\_acc\_x$ exhibit greater vibration amplitudes, while $right\_leg\_acc\_x$ shows denser oscillations with smaller amplitudes after the boundary, all of which could be features of label 13. In contrast, in Figure \ref{fig:label6}, the latter half of $right\_leg\_acc\_x$ exhibits more pronounced periodic oscillations, and $right\_leg\_acc\_y$ shows larger peaks at regular intervals, which may be indicative of label 14.

Figure \ref{fig:label5} (ground truth: 3, predicted: 15) and Figure \ref{fig:label6} (ground truth: 15) display jogging (skipping) and burpees, respectively, which share a highly similar overall trend. Both activities exhibit a relatively stable acceleration phase initially, followed by significant periodic changes after the boundary. This similarity may make it challenging for the model to differentiate the two actions. However, jogging (skipping) features more stable acceleration across all body parts in the first half, which could be a defining characteristic of label 3. On the other hand, burpees exhibit longer oscillation periods in the latter half, which might be a characteristic of label 15.

These observations suggest that subtle differences in signal amplitude, periodicity, and segmental dynamics are critical for accurate activity classification. The model may have struggled to capture these subtle variations effectively, leading to the observed misclassifications. Consistent with the analysis in Section \ref{Error_analy}, we consider adopting more advanced feature extraction methods and inertial data augmentation techniques to capture these subtle variations.

%\subsection{\B{Possible Solution: Data Augmentation}}
\subsection{Discussion: Data Augmentation}

\B{To enhance the robustness and generalization capability of existing models, we summarize a series of data augmentation techniques from \cite{hopp2024ta} for the inertial data in the WEAR \cite{bock2023wear} dataset. These techniques aim to increase the diversity of the training data while preserving the intrinsic characteristics of the sensor signals. The augmentation process consists of three main steps: coordinate permutation, axis-wise normalization, and inertial data transformation.
}

\subsubsection{\B{Coordinate Permutation}}
\B{To address the variability in wearable device orientation during physical activities, the authors of \cite{CrossHAR} propose coordinate permutation for inertial data to ensure rotational invariance. For each IMU sample, they permute the $(x, y, z)$ coordinates of the acceleration data. Given that the WEAR dataset includes acceleration data from four smartwatches (each with three axes), their method generates six permutations per sample, corresponding to the six possible orthogonal rotations in three-dimensional space: $(x, y, z)$, $(x, z, y)$, $(z, y, x)$, $(z, x, y)$, $(y, x, z)$, and $(y, z, x)$. This technique, as proposed in \cite{hopp2024ta}, enhances dataset diversity while preserving the physical meaning of the sensor data.}

\subsubsection{\B{Axis-wise Normalization}}
\B{To avoid the risk of distorting channel-specific information during normalization, the authors of \cite{CrossHAR} normalize each sensor axis independently. This approach ensures that the unique characteristics of each axis are preserved, as global normalization over all data may inadvertently suppress important channel-specific features. By normalizing the 12 axes (four smartwatches $\times$ three axes) separately, their method retains the distinct patterns of each sensor, which is particularly important for capturing subtle variations in human activities.}

\subsubsection{\B{Inertial Data Transformation}}
\B{To further enrich the training data, we apply a set of simple yet effective transformations to the inertial signals. These transformations include:
\begin{itemize}
    \item \textbf{Downscaling}: Reducing the amplitude of the signal to simulate weaker motions.
    \item \textbf{Magnifying}: Increasing the amplitude of the signal to simulate more intense motions.
    \item \textbf{Inverting}: Reversing the sign of the signal to simulate opposite directional movements.
    \item \textbf{Adding Noise}: Introducing random noise to the signal to simulate real-world sensor inaccuracies.
\end{itemize}
}
\B{
These transformations significantly increase the volume of training data, expanding the dataset from 18 subjects to 282 augmented samples. While this augmentation process increases the training time by approximately 250\% (from 1.5 hours to 5 hours), the resulting diversity in the training data enhances the model's ability to generalize to unseen scenarios.}

\subsection{Comprehensive Analysis and Future Work}

In this section, we conduct a quantitative analysis of the model using a confusion matrix and visualize the input sensor data to analyze potential causes of prediction errors.
% We identify our model's limitations and their reasons as follows:
We summarize the underlying causes of model prediction errors as follows:
\begin{itemize}
    \item The model tends to confuse similar activities, such as stretching (shoulder) and stretching (hamstrings), which both fall under the stretching category. The similarity in acceleration data for these actions makes them difficult for the model to distinguish.
    \item The model fails to capture details near activity boundaries, leading to information loss or misclassification.
    \item There is an imbalance in the sample sizes of similar action categories, which may result in poor predictive performance for minority classes.
\end{itemize} 

For future improvements, we propose addressing the model's ability to capture subtle action variations and ensuring data balance:
\begin{itemize}
    \item Designing more sophisticated feature extraction mechanisms, such as integrating CNNs with transformers, to better capture subtle action differences.
    \item Enriching the dataset using inertial data augmentation techniques, such as altering inertial data by downscaling, magnifying, inverting, or adding noise.
\end{itemize}

\section{Conclusion}

%In this paper, we address the challenges of Human Activity Recognition (HAR) using sensor signals by enhancing the ActionFormer model. While ActionFormer has demonstrated strong performance in temporal action localization, its application to sensor-based HAR is hindered by high temporal dynamics and interdependencies between spatial and temporal features. To mitigate these issues, we propose an adaptive channel-wise enhancement module, integrating the Squeeze-and-Excitation strategy to selectively emphasize informative channels while minimizing additional computational overhead. Furthermore, we employ the swish activation function to better preserve directional information in sensor signals.

\B{Human Activity Recognition (HAR) has recently witnessed advancements with Transformer-based models. Especially, ActionFormer shows us a new perspectives for HAR in the sense that this approach give us the additional outputs which detect the border of the activities as well as the activity labels. ActionFormer is originally proposed with its input as  video. However, this was converted to with its input as sensor signals as well. 
We analyze this extensively in terms of the deep learning architectures. 
Based on the report of high temporal dynamics which limits the model's ability to capture subtle changes effectively and of the interdependencies between the spatial and temporal features. We propose the modified ActionFormer which will decrease these defects for sensor signals. The key to our approach lies in accordance with the Sequence-and-Excitation strategy to minimize the increase in additional parameters and opt for the swish activation function to retain the information about direction in the negative range. Experiments on the WEAR dataset show that our method achieves substantial improvement of a 16.01\% in terms of average mAP for inertial data.}

\B{There are several possible directions for future work. First, our CE-ActionFormer not only detects activity labels but also identifies the start and end points of activities. However, in real-world data, these start and end points are usually not marked. Our CE-ActionFormer could be deployed in this context to address this challenge.
For example, while multimodal foundation models can process IMU data, they typically lack the ability to segment raw signals. The unique characteristics of our CE-ActionFormer make it well-suited for this task.
Additionally, the architecture of our CE-ActionFormer can be adapted to fit with recent vision transformer models, particularly those that include projection layers.}

%This paper presents a Channel-wise Enhancement framework that significantly improves activity localization HAR tasks using sensor signals. By amplifying critical channel information through adaptive channel-wise enhancement module, the proposed method achieves substantial improvements on the inertial data of WEAR dataset compared with ActionFormer as baseline. Furthermore, we conduct ablation studies to investigate the internal architecture of the modules. Moreover, taking into account the differences between sensor signal and video data, we propose maxpool channel-wise enhancement module to adapt this situation. 

%We analyze potential causes of model errors through numerical analysis of results and visualization of signal patterns, future work will explore advanced feature extraction techniques and data augmentation to further enhance model performance.

\bibliographystyle{plain}
\bibliography{bibtex}

\appendix
\section{Additional Results for ActionFormer for Vision}
%\section{\B{Appendix}}
\B{In the appendix, we describe additional visual results. In addition to testing our method on the visual data from the WEAR dataset \cite{bock2023wear}, we also conduct tests on several widely used video datasets. Specifically, we used I3D features extracted from the THUMOS14 \cite{idrees2017thumos} dataset, features extracted using both I3D \cite{i3d} and TSP \cite{tsp} methods from the ActivityNet-1.3 \cite{caba2015activitynet} dataset, and features from the EPIC-Kitchens 100 \cite{dima2020rescaling} dataset, where specific actions are described as verbs and objects as nouns. The baseline selection and naming convention are consistent with those used in the inertial experiments.
As shown in Table \ref{tab:visual}, our method outperforms the baseline models across all datasets. Notably, we notice that the module using adaptive average pool layer
performs well not only on the inertial data but also on the visual data from the WEAR \cite{bock2023wear}. Therefore, we also include AF-AdapAvgPool in the video feature tests. On the WEAR \cite{bock2023wear} dataset, our method achieved an average mAP of 62.24\% at tIoU thresholds
ranging from 0.3 to 0.7 with 5 steps, surpassing the baseline ActionFormer by 7.8\%. With a tIoU threshold of 0.7, our method surpasses the baseline by 24.06\%
On the I3D features from the THUMOS14 \cite{idrees2017thumos} dataset, our AF-MaxPool improve the average mAP by 0.54\% over the baseline. On other datasets, CE-HAR also lead to improvements in the performance of the original models to varying degrees.}

\begin{table*}[!ht]
\caption{\B{Results of the baseline and our improved models with 
% the CE-HAR
our method on visual data across a variety of challenging datasets, namely WEAR \cite{bock2023wear}, THUMOS14 \cite{idrees2017thumos}, ActivityNet-1.3 \cite{caba2015activitynet}, EPIC-Kitchens 100 \cite{dima2020rescaling}. The models are evaluated using mean Average Precision (mAP) across various temporal intersection over union (tIoU) thresholds when clip lengths (CL) is set to 0.5. The best results are highlighted in bold.}}
\label{tab:visual}
\resizebox{\linewidth}{!}{
\begin{tabular}{cccccccc}
Dataset                                                                              & Model                                                              & \multicolumn{6}{c}{mAP}                                                                                                                                                                                                          \\ \cline{3-8} 
                                                                                     &                                                                    & 0.3                                    & 0.4                                    & 0.5                                    & 0.6                           & 0.7                                   & Avg                           \\ \hline
                                                                                     & \cellcolor[HTML]{C0C0C0}Baseline (ActionFormer {[}Zhang et al.{]}) & \cellcolor[HTML]{C0C0C0}67.80          & \cellcolor[HTML]{C0C0C0}64.97          & \cellcolor[HTML]{C0C0C0}59.14          & \cellcolor[HTML]{C0C0C0}48.65 & \cellcolor[HTML]{C0C0C0}31.66         & \cellcolor[HTML]{C0C0C0}54.44 \\
                                                                                     & AF-MaxPool (Ours)                                                    & \textbf{69.53}                         & \textbf{67.95}                         & \textbf{63.65}                         & 57.24                         & 44.85                                 & 60.64                         \\
\multirow{-3}{*}{\begin{tabular}[c]{@{}c@{}}WEAR\\ (i3d)\end{tabular}}               & AF-AdapAvgPool (Ours)                                                   & 67.77                                  & 65.49                                  & 63.21                                  & \textbf{59.01}                & \textbf{55.72}                        & \textbf{62.24}                \\ \hline
                                                                                     & \cellcolor[HTML]{C0C0C0}Baseline (ActionFormer {[}Zhang et al.{]}) & \cellcolor[HTML]{C0C0C0}\textbf{82.21} & \cellcolor[HTML]{C0C0C0}77.81          & \cellcolor[HTML]{C0C0C0}70.29          & \cellcolor[HTML]{C0C0C0}57.70 & \cellcolor[HTML]{C0C0C0}43.96         & \cellcolor[HTML]{C0C0C0}66.39 \\
\multirow{-2}{*}{\begin{tabular}[c]{@{}c@{}}THUMOS14\\ (i3d)\end{tabular}}           & AF-MaxPool (Ours)                                                    & 81.85                                  & \textbf{77.94}                         & \textbf{70.74}                         & \textbf{59.61}                & \textbf{44.53}                        & \textbf{66.93}                \\ \hline
                                                                                     & \cellcolor[HTML]{C0C0C0}Baseline (ActionFormer {[}Zhang et al.{]}) & \cellcolor[HTML]{C0C0C0}64.27          & \cellcolor[HTML]{C0C0C0}\textbf{59.60} & \cellcolor[HTML]{C0C0C0}54.73          & \cellcolor[HTML]{C0C0C0}48.33 & \cellcolor[HTML]{C0C0C0}41.84         & \cellcolor[HTML]{C0C0C0}53.75 \\
\multirow{-2}{*}{\begin{tabular}[c]{@{}c@{}}ActivityNet-1.3\\ (tsp)\end{tabular}}    & AF-MaxPool (Ours)                                                    & \textbf{64.40}                         & 59.57                                  & \textbf{54.82}                         & 48.33                         & \textbf{41.89}                        & \textbf{53.80}                \\ \hline
                                                                                     & \cellcolor[HTML]{C0C0C0}Baseline (ActionFormer {[}Zhang et al.{]}) & \cellcolor[HTML]{C0C0C0}64.31          & \cellcolor[HTML]{C0C0C0}59.53          & \cellcolor[HTML]{C0C0C0}54.28          & \cellcolor[HTML]{C0C0C0}47.91 & \cellcolor[HTML]{C0C0C0}40.90         & \cellcolor[HTML]{C0C0C0}53.39 \\
\multirow{-2}{*}{\begin{tabular}[c]{@{}c@{}}ActivityNet-1.3\\ (i3d)\end{tabular}}    & AF-MaxPool (Ours)                                                    & \textbf{64.41}                         & \textbf{59.54}                         & \textbf{54.42}                         & \textbf{47.98}                & \textbf{41.19}                        & \textbf{53.51}                \\ \hline
                                                                                     & \cellcolor[HTML]{C0C0C0}Baseline (ActionFormer {[}Zhang et al.{]}) & \cellcolor[HTML]{C0C0C0}24.23          & \cellcolor[HTML]{C0C0C0}22.11          & \cellcolor[HTML]{C0C0C0}19.05          & \cellcolor[HTML]{C0C0C0}13.95 & \cellcolor[HTML]{C0C0C0}8.86          & \cellcolor[HTML]{C0C0C0}17.64 \\
\multirow{-2}{*}{\begin{tabular}[c]{@{}c@{}}EPIC-Kitchens 100\\ (verb)\end{tabular}} & AF-MaxPool (Ours)                                                    & \textbf{25.31}                         & \textbf{22.22}                         & \textbf{19.47}                         & \textbf{14.33}                & \textbf{9.07}                         & \textbf{18.08}                \\ \hline
                                                                                     & \cellcolor[HTML]{C0C0C0}Baseline (ActionFormer {[}Zhang et al.{]}) & \cellcolor[HTML]{C0C0C0}21.93          & \cellcolor[HTML]{C0C0C0}19.92          & \cellcolor[HTML]{C0C0C0}\textbf{16.56} & \cellcolor[HTML]{C0C0C0}11.62 & \cellcolor[HTML]{C0C0C0}\textbf{7.29} & \cellcolor[HTML]{C0C0C0}15.46 \\
\multirow{-2}{*}{\begin{tabular}[c]{@{}c@{}}EPIC-Kitchen 100\\ (noun)\end{tabular}}  & AF-MaxPool (Ours)                                                    & \textbf{22.26}                         & \textbf{19.97}                         & 16.41                                  & \textbf{12.44}                & 6.98                                  & \textbf{15.61}                \\ \hline
\end{tabular}}
\end{table*}

%\newpage
%\bibliographystyle{plain}
%\bibliography{bibtex}

\end{document}